\RequirePackage{fix-cm}
\documentclass[twocolumn]{svjour3}          
\smartqed  
\usepackage{graphicx}
\usepackage{natbib}
\usepackage{amsmath}
\usepackage{amssymb}
\usepackage{bbm}
\usepackage{bm}
\usepackage{booktabs}
\usepackage{array} 
\usepackage{blindtext}
\usepackage{comment}
\usepackage[table]{xcolor}

\newcommand{\major}[1]{{\color{black}{#1}}} 
\newcommand{\minor}[1]{{\color{black}{#1}}} 

\usepackage{pifont}

\usepackage[dvipsnames]{xcolor}
\usepackage{color, colortbl}
\definecolor{citecolor}{HTML}{0071bc}
\definecolor{tabhighlight}{HTML}{e5e5e5}
\usepackage[colorlinks,citecolor=citecolor]{hyperref}

\makeatletter
\renewcommand\paragraph{
  \@startsection{paragraph} 
  {4} 
  {\z@} 
  {.5em \@plus1ex \@minus.2ex} 
  {-.5em} 
  {\normalfont\normalsize\bfseries} 
}
\makeatother
\usepackage{algorithm}
\usepackage{amsfonts}
\usepackage[caption=false,font=normalsize,labelfont=sf,textfont=sf]{subfig}
\usepackage{textcomp}
\usepackage{stfloats}
\usepackage{url}
\usepackage{verbatim}
\usepackage{cite}
\usepackage{epsfig}
\usepackage{ifthen}
\usepackage{caption}
\usepackage{multicol}
\usepackage{multirow}
\usepackage{makecell}
\usepackage{arydshln}
\usepackage{mathtools}
\usepackage{algpseudocode}
\usepackage{xfrac}
\usepackage{setspace}

\usepackage{subcaption}
\usepackage{booktabs}
\usepackage{graphicx}
\usepackage{algorithm}
\usepackage{enumitem}

\newcommand{\ie}{\textit{i.e.}}
\newcommand{\eg}{\textit{e.g.}}
\newcommand{\etal}{\textit{et~al.}}

\begin{document}
\sloppy

\title{AITTI: Learning Adaptive Inclusive Token for \\ Text-to-Image Generation 
}


\author{Xinyu Hou   \and
        Xiaoming Li \and
        Chen Change Loy
}


\institute{Xinyu Hou \at
              S-Lab, Nanyang Technological University, Singapore \\
              \email{xinyu.hou@ntu.edu.sg}
           \and
           Xiaoming Li \at
              S-Lab, Nanyang Technological University, Singapore \\
              \email{csxmli@gmail.com}
           \and
           Chen Change Loy (Corresponding author) \at
              S-Lab, Nanyang Technological University, Singapore \\
              \email{ccloy@ntu.edu.sg}
}

\date{Received: date / Accepted: date}

\maketitle

\begin{abstract}
Despite the high-quality results of text-to-image generation, stereotypical biases have been spotted in their generated content, compromising the fairness of generative models. 
In this work, we propose to learn \textit{adaptive inclusive tokens} to shift the attribute distribution of the final generative outputs. 
Unlike existing debiasing methods, our approach operates without explicit attribute \minor{class} specification \minor{during inference} or prior knowledge of \minor{the original} bias distribution. 
Specifically, the core of our method is a lightweight \textit{adaptive mapping network}, which can customize the inclusive tokens for the concepts to be de-biased, making the tokens \textit{generalizable to unseen concepts} regardless of their original bias distributions.
This is achieved by tuning the adaptive mapping network with a handful of balanced and inclusive samples using an anchor loss.
Experimental results demonstrate that our method outperforms previous bias mitigation methods without attribute specification while preserving the alignment between generative results and text descriptions. Moreover, our method achieves comparable performance to models that require specific attribute \minor{classes} or editing directions for generation. Extensive experiments showcase the effectiveness of our adaptive inclusive tokens in mitigating stereotypical bias in text-to-image generation. The code is publicly available at \url{https://github.com/itsmag11/AITTI}.
\end{abstract}

\section{Introduction}
\label{sec:introduction}

Text-to-image (T2I) generation has gained widespread popularity due to its ability to create visual content from user-defined text descriptions. However, alongside these advancements, concerns have emerged regarding the presence of stereotypical biases in the generated outputs, as highlighted by various studies~\citep{ghosh2023person, chinchure2023tibet, wang2023t2iat, bianchi2023easily, wang2024new, jha2024surface}. 

Bias in T2I generation is often manifested in the unequal representation of social groups. Specifically, when no explicit attribute \minor{classes} are specified in a prompt, T2I models tend to generate human figures that resemble certain genders or races, inadvertently reinforcing harmful social stereotypes~\citep{ghosh2023person, bianchi2023easily}. Certain occupations, for example, are strongly associated with specific genders, such as the stereotype of doctors being male and nurses being female. Moreover, prompts related to negative concepts (\eg, poverty or unattractiveness) disproportionately generate images of individuals from underrepresented racial groups~\citep{bianchi2023easily, ghosh2023person, jha2024surface}. This unfair representation can perpetuate stereotypes, justify societal discrimination, and undermine the rights of minority groups to be treated equitably. 

While biased training datasets are often identified as the source of these biases, the pervasive nature of societal stereotypes complicates efforts to eliminate them entirely. 
\minor{Evaluations across multiple T2I backbones in Tab.~\ref{tab:basemodels} further confirm that such biases not only persist but in some cases intensify with scaling. For instance, SD3 and SDXL exhibit stronger gender and race disparities than SD1.5, despite being trained on substantially larger datasets. These findings suggest that increasing model capacity and training data do not resolve fairness issues. In fact, scaling may amplify existing imbalances in the corpus. This underscores the necessity of dedicated debiasing methods, as stronger generative capacity does not automatically entail more inclusive generation.}

\begin{table*}[!t]
    \centering
    \caption{\major{Biases in Text-to-Image models. $D_{KL}$ metric measures the KL divergence between the generated attribute distribution and the ideal one. Smaller values indicate a closer match to the ideal distribution. Details about the evaluation protocols are explained in Sec.~\ref{sec:experiments}.}}
    \begin{tabular}{c|c|c|c}
    \hline
       Model & Gender $D_{KL} \downarrow$ & Race $D_{KL} \downarrow$ & Age $D_{KL} \downarrow$ \\
       \hline
       SD1.5~\citep{Rombach2022stable} & \textbf{0.3584} & 0.5973 & 0.2319 \\
       SD2.1~\citep{Rombach2022stable} & 0.4166 & \textbf{0.5819} & 0.2062 \\
       SDXL~\citep{podell2023sdxl} & 0.4919 & 0.6796 & \textbf{0.1965} \\
       SD3~\citep{sd3} & 0.5480 & 0.6770 & 0.4319 \\
       FLUX~\citep{flux_github} & 0.5354 & 0.5916 & 0.2868 \\
       \hline
    \end{tabular} 
    \label{tab:basemodels}
\end{table*}

\begin{figure}
    \centering
    \includegraphics[width=\linewidth]{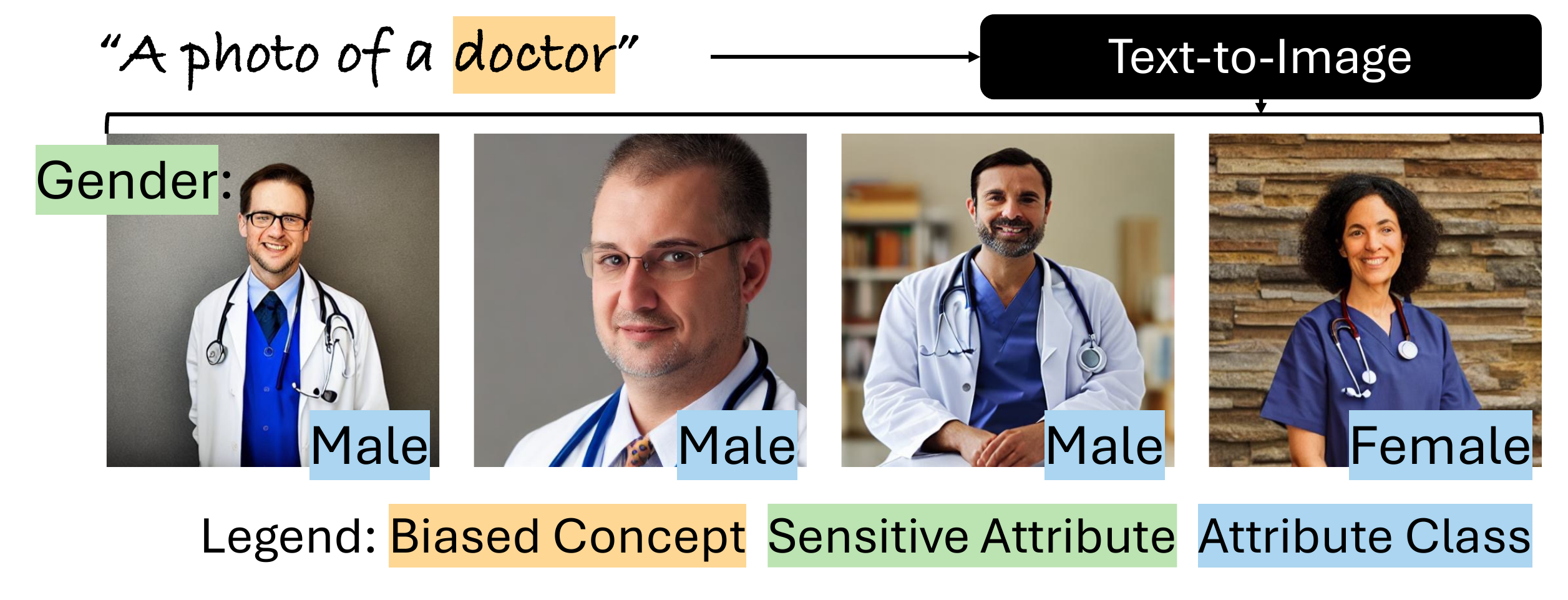}
    \caption{\textcolor{black}{Illustration of key concepts in T2I bias mitigation. The input prompt may contain biased concepts (\eg, doctor), and the generated images predominantly depict some attribute \minor{classes} (\eg, male) over others, reflecting bias with respect to a sensitive attribute (\eg, gender). Such imbalances highlight the need for fairness-aware generation techniques.}}
    \label{fig:con}
\end{figure}

The goal of this paper is to develop a bias mitigation method that addresses biased concepts. We define key terms as follows: \textit{sensitive attributes} $\mathcal{A}$ refer to attributes of interest for fairness (\eg, gender), and \textit{attribute classes} $\mathcal{A}_c$ represent the possible classes of the attribute (\eg, female and male). \textit{Biased concepts} $\mathcal{C}$ refer to concepts that may carry stereotypes (\eg, doctor). Figure~\ref{fig:con} illustrates the notion of different terms.

An ideal inclusive T2I model would generate outputs with evenly distributed sensitive attributes across all attribute classes when no \minor{attribute-class-related} instructions are provided. Specifically, a critical feature of a fair model is its ability to produce inclusive outcomes without explicit instructions regarding the target attribute class. Furthermore, users' potential unawareness of biases related to a concept should be respected. Therefore, we believe that, for a neutral concept, a good debiasing algorithm should
\begin{enumerate}
    \item Generate fairer results \textit{without} explicit specification of the target attribute class \minor{during inference}, and
    \item Require \textit{no prior knowledge} of the original bias distribution associated with the concept, \eg, the doctor concept being stereotypically biased toward males.
\end{enumerate}
In practice, we hope that by specifying the sensitive attribute of interest rather than the specific attribute class \minor{or editing direction} in the prompt (\eg, ``A $\mathord{<}$gender-inclusive$\mathord{>}$ doctor'' instead of ``A female doctor''), users can obtain fair generative results more seamlessly for the attribute they care about.

Achieving the aforementioned inclusive properties is challenging in the absence of direct attribute specification and prior knowledge about the bias distribution. \minor{Here, ``attribute specification'' refers specifically to attribute class or bias direction specification during generation, rather than the type of sensitive attribute itself.} Our study demonstrates that the simple approach of learning a fixed inclusive token via Textual Inversion (TI)\citep{gal2022textual} is ineffective in mitigating biases across concepts with varying prior distributions (see Sec.~\ref{sec:ti} for a detailed discussion). Moreover, this approach risks altering the semantic meaning of the original prompt. 

In this work, we adopt the mainstream prompt-tuning method but focus on learning \textbf{adaptive inclusive tokens} that do not specify any target class, yet can shift the biased attribute in the generated output toward a more equitable distribution, irrespective of the class to which it was originally biased.
We hypothesize that the token embeddings of a biased concept (\eg, the doctor concept) encode information about its bias distribution. To enable the adaptability of the learned inclusive tokens across different biased concepts, we introduce a lightweight \textbf{adaptive mapping network} that finds the optimal inclusive token, tailored to the target concept. Furthermore, we propose an \textbf{anchor loss} to guide the desired properties of the inclusive token. This loss minimizes the discrepancy between denoising UNet predictions of a prompt with the inclusive token and the prompt with the ground truth attribute class. The anchor loss ensures that the inclusive token has an aligned impact with attribute class tokens, enabling alternation among all possible classes.

Our method is validated on the widely adopted Stable Diffusion (SD) framework~\citep{Rombach2022stable}. By using small-scale balanced datasets generated by the SD model itself to train our adaptive inclusive tokens, we demonstrate a significant improvement in the fairness of the model outputs. Notably, the learned adaptive inclusive tokens exhibit generalizability to unseen concepts and prompts, and can be concatenated to mitigate multiple biases across various attributes.

The contributions of this work are as follows. We propose a simple yet effective prompt-tuning approach for inclusive text-to-image generation without the need for attribute \minor{class} specification or prior knowledge of  \minor{the original bias distribution of the} biased concepts. Specifically, we introduce the adaptive mapping network and anchor loss to address the issue of generalizability across different attribute class dominations. Extensive experiments validate the effectiveness of our method both quantitatively and qualitatively.

\section{Related Work}
\label{sec:related_work}

\subsection{Bias in Text-to-Image Generation} 
Comprehensive analyses have been conducted to study the bias and unfairness observed in T2I generation~\citep{bianchi2023easily, chinchure2023tibet, ghosh2023person, jha2024surface, wang2023t2iat, wang2024new, smiling, lyu2025existing, teo2023measuring, openbias}. \citet{wang2023t2iat} first introduce the Implicit Association Test (IAT)~\citep{Greenwald1998IAT} from social psychology to measure biases in the task of T2I. IAT is designed to reveal implicit biases that an individual may hold unconsciously towards certain concepts. By experimenting with the valence and stereotype IATs on T2I output images, it is found that valence and stereotypical biases exist in state-of-the-art T2I models at various scales, \eg, the pleasant attitude is significantly biased towards straight sexual orientation than homosexual ones. \citet{bianchi2023easily} study a wide range of stereotypes related to gender, race, nationality, and other identities associated with traits, occupation, and even object descriptions. They conclude that attempts to either specify counter-stereotype prompts by users or add system ``guardrails'' cannot prevent stereotypes from spreading in the T2I results. \citet{ghosh2023person} spot the over-representation of Caucasian males in general terms like ``person'' and the over-sexualization of women of color without specification. TIBET~\citep{chinchure2023tibet} proposes to identify and measure biases in any T2I models using counterfactual reasoning, which breaks the limitation of pre-defined bias axes in previous studies. \citet{wang2024new} propose BiasPainter, which provides an automatic and systematic study of gender, race, and age biases. It augments a seed image of a clear identity with queries of professions, activities, types of objects, and personality traits, and compares the attributes of the identity between augmented and seed images to identify biases in queries. \citet{jha2024surface} focus on exploring geo-cultural stereotypes in T2I models with a large scale of nationality-based identity groups. It is revealed that the severity of bias varies for different identity groups, and the ``default'' representations of identity groups contain stereotypical appearances.

\subsection{Bias Mitigation in Text-to-Image Generation} 
Various approaches have been developed to alleviate stereotypical biases in T2I generation, including model fine-tuning~\citep{runwayresearch2023mitigating, kim2023destereotyping, shen2023finetuning, zhou2024association, jiang2024debiasdiff}, prompt tuning~\citep{gal2022textual, bansal2022howwell, zhang2023itigen, teo2024fairqueue, li2023fairmapping}, concept editing~\citep{orgad2023time, gandikota2023unified, yesiltepe2024mist}, and inference guidance~\citep{friedrich2023fairdiffusion, parihar2024balancing, choi2024switchmechanism}. 
A straightforward strategy is to fine-tune the entire T2I model on a large-scale dataset that is carefully synthesized to cover various classes of bias attributes~\citep{runwayresearch2023mitigating}. Besides, techniques of fine-tuning the sampling process of the diffusion model have been proposed so that distributional constraints on the generative outputs, which are the direct interpretation of fairness, can be applied~\citep{kim2023destereotyping, shen2023finetuning}. However, such fine-tuning methods require a heavy load of computations. 
\citet{jiang2024debiasdiff} instead fine-tune a few adapters, each leads to images with one class. Together with a distribution indicator and a gating function, fair results across all classes are achieved. To mitigate bias arising due to the association of objects, \citet{zhou2024association} fine-tune the diffusion model to take in sensitive constraints predicted by a pre-trained Prompt-Image-Stereotype CLIP and a Sensitive Transformer.
Prompt tuning-based methods aim to modify or add textual tokens to affect the T2I generation outputs. \citet{gal2022textual} learn a fairer word for a biased concept from a small dataset. \citet{bansal2022howwell} explore the impact of ethical interventions added to the original prompts on the fairness of generative results. \citet{li2023fairmapping} introduce a fair mapping network that projects the text embeddings of a neutral prompt to the middle of prompts with all possible classes. The above prompt tuning methods do not modify the prompt to provide explicit guidance during generation. On the contrary, \citet{zhang2023itigen} propose to overfit one class token for each attribute class by image prompts and apply ad-hoc post-processing to loop over all combinations of target attribute \minor{classes} to achieve inclusive generation. Then, \citet{teo2024fairqueue} further refine ITI-GEN~\citep{zhang2023itigen}'s quality degradation by prompt queuing and attention amplification.
\citet{orgad2023time}, \citet{gandikota2023unified}, and \citet{yesiltepe2024mist} apply model editing to enforce the generation of non-stereotypical classes by optimizing the cross-attention weights of the diffusion models. However, determining the appropriate editing strengths is challenging, considering the large variations in bias strengths across different concepts and classes.
Another line of retraining-free approaches involves incorporating desired guidance during inference. \citet{friedrich2023fairdiffusion} apply fair guidance at inference by steering biased concepts in predefined directions to enhance fairness. \citet{choi2024switchmechanism} switch the generation guidance from one attribute class to another at the midpoint of denoising. While these methods avoid the need for training or fine-tuning, it relies on prior knowledge of the biased concept and requires each generation to be guided by a specific semantic direction. Similarly, \citet{parihar2024balancing} employ a pre-trained h-space classifier to provide explicit distribution guidance during inference. Although this method does not require retraining the diffusion model, it demands additional effort to train the classifier and increases the computational cost of inference due to classifier-guided optimization.
Our approach adopts the prompt tuning approach without necessitating heavy model fine-tuning or prior knowledge about the original bias distribution, making it computationally efficient and highly generalizable.

\subsection{Prompt Tuning} 
Prompt tuning is a technique that adapts a large language model (LLM) to new concepts by optimizing some prompt parameters with the model weights fixed. Prompt tuning has been applied to various downstream tasks such as image classification~\citep{zhou2022coop, zhou2022cocoop}, customized generation~\citep{gal2022textual, ruiz2022dreambooth}, and bias mitigation~\citep{gal2022textual, kim2023destereotyping}. Zhou \etal~\citep{zhou2022coop, zhou2022cocoop} propose to optimize trainable context tokens in class prompts to boost the performance of zero-shot classification using CLIP classifier~\citep{radford2021clip} and further tailor the tokens to be input-conditioned for better generalization to unseen classes. Textual Inversion~\citep{gal2022textual} is proposed to invert a visual concept from a small set of images to a new pseudo word and customize generation on the visual concept. Similarly, \citet{ruiz2022dreambooth} and \citet{kumari2022customdiffusion} update a trainable token together with some parts of the model to represent a particular visual content. The ability of prompt tuning to mitigate biases has been demonstrated by \citet{gal2022textual} as mentioned in the previous subsection. \citet{kim2023destereotyping} adopt a similar prompt tuning approach. However, they optimize the tokens on the sampling stage of the diffusion model, leading to an extensive amount of computation and memory required.
Our method tackles the limitation of fixed inclusive tokens on the transferability to different domination classes, as observed in both previous methods, by learning adaptive inclusive tokens.

\section{Methodology}
\label{sec:methodology}

\subsection{Preliminaries}
\label{sec:preliminary}

\noindent
\textbf{Diffusion Model.} Stable Diffusion~\citep{Rombach2022stable} is a commonly used latent diffusion model for image generation. With classifier-free guidance~\citep{ho2021classifier} of textual conditions, SD demonstrates excellent performance in the T2I generation task. During each training step, a training image $x_0$ is first encoded into a latent space as $z_0$ by a pre-trained image encoder $\mathcal{E}(\cdot)$. Then, a noise latent and a denoising timestep $t$ are sampled to compute the ground truth noise $\epsilon$, which will be added to the encoded image to obtain the noisy latent $z_t$ in the current timestep. On the other side, a textual prompt $T$ that describes the image is tokenized to token embeddings $v_T$ and encoded to textual embeddings $e_T$ by the pre-trained CLIP text encoder~\citep{radford2021clip}. Subsequently, a denoising UNet $\epsilon_\theta$~\citep{unet} takes in the noisy image latent, the denoising timestep, and the textual condition to predict the noise added to the image latent. The learning objective of the denoising stage is formulated as follows:

\begin{equation}
        \hspace{0.2cm} 
        \mathcal{L}_\textit{denoise}=\mathbb{E}_{z_0, \epsilon\sim\mathcal{N}(0,1), t, e_\textit{T}}\left[\left\|\epsilon-\epsilon_\theta\left(z_t, t, e_T\right)\right\|_2^2\right]\ .
    \label{eqn:ldm}
\end{equation}
\noindent
During inference, the denoising UNet predicts the noise to be removed at each timestep from a Gaussian noise latent, conditioned on textual input.

\label{sec:ti}
\noindent \textbf{Is Textual Inversion Effective for Debiasing?}
Our method is inspired by the Textual Inversion framework introduced by \citet{gal2022textual}. The authors briefly explored the potential of textual inversion for bias reduction with a simple demonstration. Specifically, they curated a small, diverse dataset for a specific concept, \eg, doctor, and learned a fairer word to replace the original concept in text prompts. Their results suggest that fairness in T2I models can be significantly influenced by textual conditions, and that it is feasible to learn a pseudo word from a small set of balanced images to promote a more equitable distribution.

The original Textual Inversion framework is concept-specific, \ie, it learns a pseudo word \textcolor{black}{$\mathord{<}i\mathord{>}$ representing} $\mathord{<}$inclusive-doctor$\mathord{>}$ to replace ``doctor'', and the pseudo word cannot be reused for other concepts.
To construct a stronger baseline that could generalize to unseen concepts, we revise their approach by disentangling concept-specific information from the learned tokens, ensuring that they solely encode the inclusiveness of a biased attribute. \textcolor{black}{In particular, we revise the original training prompt template by adding the concept noun of the training sample after the pseudo word, \eg ``a photo of a $\mathord{<}i\mathord{>}$'' becomes ``a photo of a $\mathord{<}i\mathord{>}$ doctor'', so that $\mathord{<}i\mathord{>}$ now represents} $\mathord{<}$gender-inclusive$\mathord{>}$ that can be applied to any human-related concept.

The quantitative results of applying this naïve revision (denoted as rTI) on SD1.5 are reported in Tab.~\ref{tab:chief_comparisons-a}. In this experiment, we tested the approach on ten unseen occupations\footnote{Male-dominated (MD): \texttt{["doctor", "chief", "farmer", "architect", "software developer"]}, Female-dominated (FD): \texttt{["ballet dancer", "yoga instructor", "cosmetologist", "fashion designer", "flight attendant"]}} that are stereotypically associated with different genders (\eg, ``software developer'' is male-dominated, while ``flight attendant'' is female-dominated).
Although \textcolor{black}{rTI} numerically reduces biases in gender and race attributes, as observed from the reduced distribution discrepancy~\citep{zhang2023itigen} (lower is better, more details in Sec.~\ref{sec:setup}), it has two key limitations in its generated outputs. 
\begin{enumerate}
    \item A single learned inclusive token is insufficient for mitigating biases in concepts originally skewed toward different attribute classes.
    \item The learned inclusive token can unintentionally alter the semantic meaning of certain concepts.
\end{enumerate}

\begin{table}[t]
    \centering
    \renewcommand\arraystretch{1.2}
    \caption{
    Revised Textual Inversion (rTI) uses a fixed inclusive token for debiasing. Although it can now handle unseen concepts, the effect of debiasing is not satisfactory, given its uneven influences across occupations stereotypically dominated by different genders.  We use distribution discrepancy $\mathcal{D}_{KL}$ as the quantitative metric, lower is better~\citep{zhang2023itigen}. Each domination group consists of five unseen occupations. \textbf{MD}: male-dominated occupations. \textbf{FD}: female-dominated occupations.
    }
    \setlength{\tabcolsep}{3.56mm}
	{
    \begin{tabular}{c|cc}
    \hline
    \textbf{Method} & \textbf{MD} $\downarrow$ & \textbf{FD} $\downarrow$ \\ 
    \hline
    SD1.5 wihout rTI & 0.5166 & 0.4401 \\ 
    SD1.5 with rTI & \major{0.1862} & \major{0.5480} \\ 
    \hline
    \end{tabular}
    }
    \label{tab:chief_comparisons-a}
\end{table}

As shown in Tab.~\ref{tab:chief_comparisons-a}, 
\major{the revised Textual Inversion (rTI) improves fairness in male-dominated occupations, but even increases bias for female-dominated ones when using a single fixed inclusive token. This gap in performance on different domination groups suggests that the fixed token only learn to shift results toward generating more female figures rather than learning a fair distribution. That is, more females for ``doctor'' but also more females for ``yoga instructor''.}
This observation aligns with the findings of \citet{kim2023destereotyping}, who reported that their de-stereotyping prompt had limited transferability to unseen classes that do not share the same dominant attribute class.

\begin{figure}[!t]
    \centering
    \includegraphics[width=\linewidth]{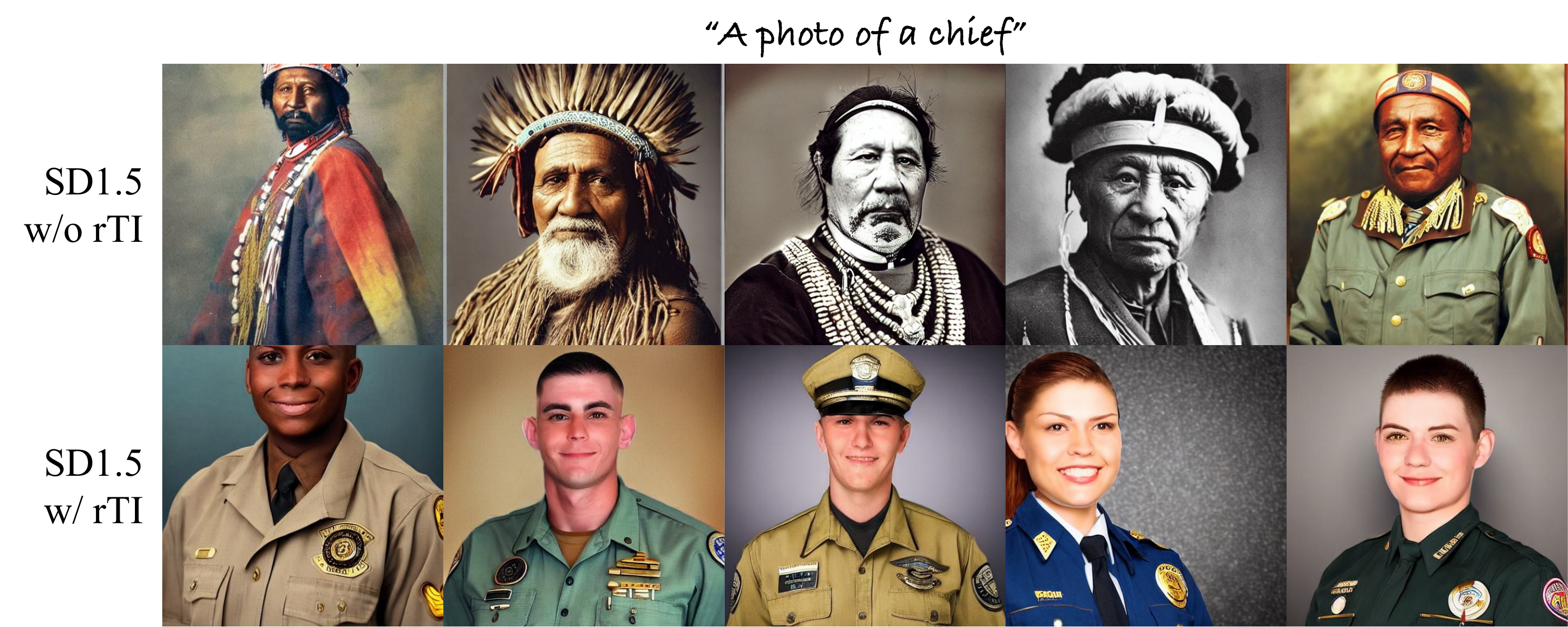}
    \caption{Revised Textual Inversion (rTI) with fixed inclusive token causes semantic drifting of visual concepts. All images are generated with the same random seed. The caption above indicates the base prompt $T(c)$. \textbf{Top}: SD1.5 without rTI; \textbf{Bottom}: SD1.5 with rTI.}
    \label{fig:chief_comparisons-b}
\end{figure}

Moreover, as illustrated in Fig.~\ref{fig:chief_comparisons-b}, the term ``chief'' exhibits semantic drift under the influence of the inclusive token learned via the revised Textual Inversion. While the modified representations are not inherently incorrect—given the broad definition of ``chief''—our goal is to ensure that the learned inclusive tokens affect only the biased attributes of the generated human figures without altering the semantics of the underlying concept.

\begin{figure*}[!t]
    \centering
    \includegraphics[width=\textwidth]{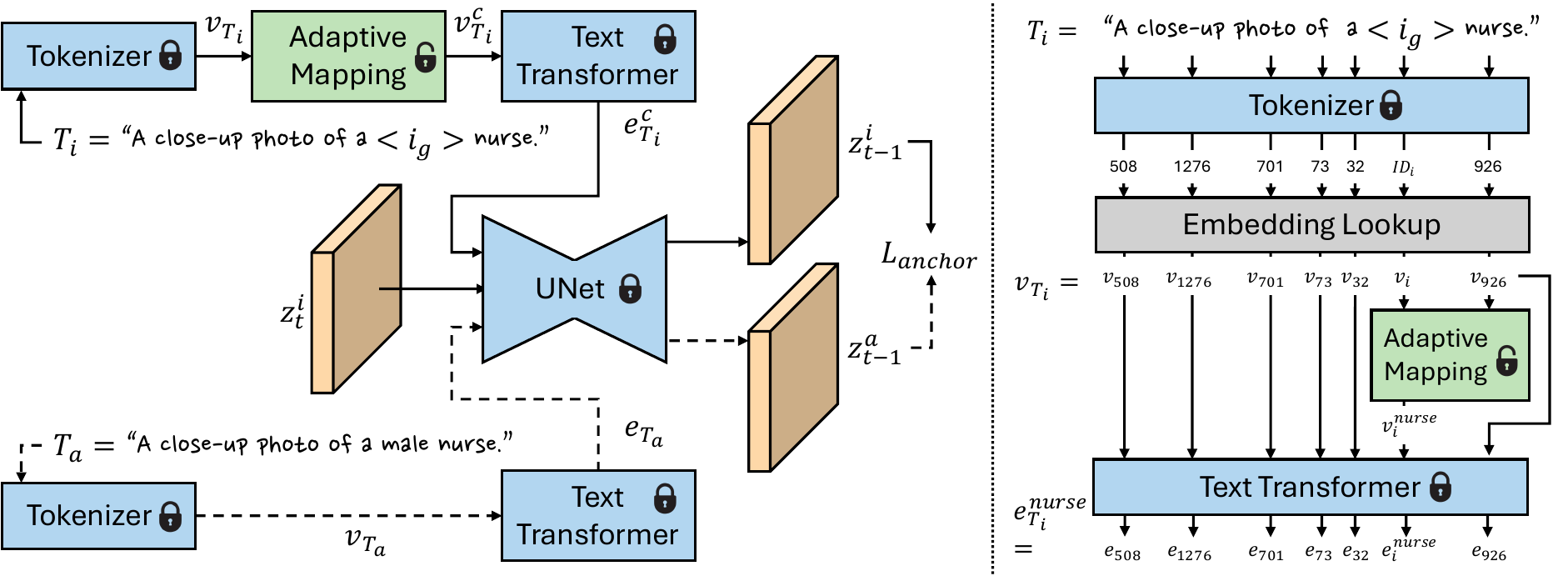}
    \caption{Framework of our proposed adaptive inclusive token for text-to-image generation. The blue color indicates frozen weights, and the green color indicates trainable weights. \textbf{Left}: single training stage. \textbf{Right}: details of the text model with the adaptive mapping network. The adaptive inclusive token is concept-specific. $TokenIDs$ are for illustration only.}
    \label{fig:framework}
\end{figure*}

\subsection{Adaptive Inclusive Token}
\label{sec:ait}

To address the limitations of a fixed inclusive token \textit{across all concepts}, we propose a lightweight adaptive mapping network that dynamically adjusts the token based on the target concept. To prevent the learned token from capturing irrelevant information from the training set, which could potentially cause semantic drift in biased concepts and degrade fairness in the generated results, we introduce an anchor loss to constrain its effect. The overall framework of our method is illustrated in Fig.~\ref{fig:framework}. Below, we detail the key components of our approach.

\noindent \textbf{Adaptive Mapping Network.} As shown on the right side of Fig.~\ref{fig:framework}, the adaptive mapping network $\mathcal{F}_{am}$ is incorporated after the token embedding lookup and before the text transformer in our framework. The inclusive prompt $T_i$ contains an inclusive token, denoted by a placeholder $\mathord{<}i\mathord{>}$ $\in \{$$\mathord{<}i_g\mathord{>}$,$\mathord{<}i_r\mathord{>}$,$\mathord{<}i_a\mathord{>}$$\}$, corresponding to $\mathord{<}$gender-inclusive$\mathord{>}$,$\mathord{<}$race-inclusive$\mathord{>}$,$\mathord{<}$age-inclusive$\mathord{>}$, respectively. The placeholder is processed through the text tokenizer and token embedding lookup table along with other words, yielding an initial inclusive token embedding $v_i$. Then, the inclusive token embedding $v_i$ and the biased concept embeddings $v_c$ are jointly fed into the adaptive mapping network, which predicts a concept-adaptive inclusive token $v_i^c$. \minor{Notably, the inclusive token does not encode any semantic attribute class information, but only indicates the target sensitive attribute for inclusivity.}

Our key hypothesis is that the token embeddings of a concept inherently encode information about its bias distribution. Therefore, adapting the inclusive token for each concept enables a more effective shift toward fairness, regardless of the concept's original distribution. Finally, the adaptive inclusive token $v_i^c$ substitutes $v_i$ in the original token embeddings $v_{T_i}$ to form $v_{T_i}^c$, which is then passed to the text transformer model. The complete text model process with adaptive mapping is outlined in Algorithm~\ref{alg:example}.

\begin{algorithm}
\setstretch{1.16}
\caption{Text Model with Adaptive Mapping Network}\label{alg:example}
\begin{algorithmic}[1]
\State \textbf{Input:} Tokenizer $\mathcal{F}_{tokenizer}$, Token lookup table $\mathcal{L}$, Text transformer $\mathcal{F}_{text}$
\State \textbf{Input:} Inclusive prompt $T_i = [T(c); \mathord{<}i\mathord{>}]$, where $T$ is the base prompt with biased concept $c$ and $\mathord{<}i\mathord{>} \in \{$$\mathord{<}i_g\mathord{>}$,$\mathord{<}i_r\mathord{>}$,$\mathord{<}i_a\mathord{>}$$\}$ is the inclusive placeholder
\State \textbf{Input:} Adaptive Mapping Network $\mathcal{F}_{am}$
\State $Token IDs = \mathcal{F}_{tokenizer}(T_i)$
\State Token embeddings: $v_{T_i} = \mathcal{L}(Token IDs) = [v_T; v_i]$
\State Concept-adaptive inclusive token embedding: $v_i^c = \mathcal{F}_{am}(v_i, v_c)$
\State Update $v_{T_i}$ to $v_{T_i}^c = [v_T; v_i^c]$
\State Text embedding: $e_{T_i}^c = \mathcal{F}_{text}(v_{T_i}^c)$
\State \textbf{Return} $e_{T_i}^c$
\end{algorithmic}
\end{algorithm}

\noindent \textbf{Anchor Loss.} The concept drift observed in the revised Textual Inversion results (Sec.~\ref{sec:preliminary}) suggests that the inclusive token may capture unintended information during training. We hypothesize that this issue arises from the lack of explicit constraints guiding the token toward learning properties that promote fairness. To address this, we introduce an anchor loss, which constrains the UNet noise prediction conditioned on the inclusive text prompt, as this directly influences the final generative output. Prior work by \citet{li2023fairmapping} interprets fairness constraints as requiring the text embeddings of a debiased prompt to be equidistant from those of all class-specific prompts for a given bias attribute. However, we argue that fairness and inclusiveness should not be defined by an averaged distance across all possible classes, but rather by ensuring an equal-probability shift among those classes. Based on this principle, we propose the following anchor loss formulation:
\begin{equation}
    \begin{split}
        \hspace{0.7cm} 
        \mathcal{L}_\textit{anchor}=&\mathbb{E}_{z_0, \epsilon\sim\mathcal{N}(0,1), t, e_{T_i}^c, e_{T_a}} \\ &\left[\left\|\epsilon_\theta\left(z_t, t, e_{T_i}^c\right) -\epsilon_\theta\left(z_t, t, e_{T_a}\right)\right\|_2^2\right]\, .
    \end{split}
    \label{eqn:anchor}
\end{equation}

\noindent
Here, the notation follows Eqn.~(\ref{eqn:ldm}). The term $e_{T_i}^c$ represents the text embeddings of the inclusive prompt $T_i$, while $e_{T_a}$ denotes the text embeddings of the anchor prompt $T_a=[T(c);a]$, where the inclusive token is replaced by the ground-truth attribute class $a$ of the training sample. For example, for a training sample of a female firefighter with gender as the target bias attribute, if $T_i$ is ``a photo of a $\mathord{<}i_g\mathord{>}$ firefighter'' then $T_a$ will be ``a photo of a female firefighter''. 

Enforcing similarity between the UNet noise predictions with $e_{T_i}^c$ and $e_{T_a}$ conditions has two potential benefits. First, it ensures that the learned inclusive token influences only the biased attribute, \eg, gender, as the anchor word, while preserving the semantics of the remaining content. Second, it encourages inclusive generation, allowing the effect of the inclusive token to shift flexibly across all possible attribute classes rather than averaging their representations.

The overall training objective for our adaptive inclusive token is defined as:

\begin{equation}
    \hspace{2cm} 
    \mathcal{L} = \mathcal{L}_{denoise} + \lambda \cdot \mathcal{L}_{anchor},
\end{equation}
\noindent
where $\mathcal{L}_{denoise}$ is calculated based on the inclusive prompt $T_i$, and $\lambda$ is the weighting parameter that balances the two losses.

\section{Experiments}
\label{sec:experiments}

\subsection{Experimental Setup}
\label{sec:setup}

\noindent \textbf{Scope.} 
We evaluate the effectiveness of our adaptive inclusive token in mitigating biases across three commonly studied sensitive attributes—gender, race, and age—denoted as $\mathcal{A} \in \{gender, race, age\}$. These attributes correspond to real-world issues of sexism, racism, and ageism in generative models.
For gender bias, we consider binary classes of male and female, acknowledging the limitation of this approach in representing non-binary genders. We believe that attempting to explicitly identify non-binary appearances based on current datasets may reinforce stereotypes within these underrepresented groups. Therefore, we refrain from such classifications until carefully curated public datasets with non-binary representations become available.
For racial bias, we adopt the seven racial categories from the FairFace dataset~\citep{karkkainen2021fairface}. To simplify analysis, we merge East Asian and Southeast Asian into a single Asian category, resulting in six distinct racial groups: White, Black, Asian, Middle Eastern, Indian, and Latino Hispanic.
For age bias, we classify individuals into two broad groups: young and old. 
Our study focuses on neutral bias concepts related to human figures, particularly in occupational roles: $\mathcal{C} \in \{neutral\ occupations\}$ where neutral occupations are defined as those that remain factually correct across all attribute classes. For example, “waitress” is not considered neutral, as it inherently implies a female figure by definition.

\noindent \textbf{Training Data.} 
\label{sec:train data}
To construct the training set with known attribute classes, we utilize the SD1.5\footnote{runwayml/stable-diffusion-v1-5}~\citep{Rombach2022stable} model with prompts formatted as ``High-quality photo of a/an [attribute] \{occupation\}''. The [attribute] choices include ``male'', ``female'' for gender, ``White'', ``Black'', ``Asian'', ``Middle Eastern'', ``Indian'', ``Latino Hispanic'' for race, and ``young'', ``old'' for age as mentioned in Scope section. The \{occupation\} list is given in the Appendix Sec.~\ref{sec:occ list}. We generate \major{10} images for each attribute-occupation combination. In total, there are \major{$(24\times2\times20)$} images for gender attribute, \major{$(24\times6\times10)$} for race, and \major{$(24\times2\times10)$} for age. To ensure data quality, we apply the RetinaFace detector~\citep{deng2020retinaface} with a confidence threshold of 0.97, filtering out images without valid facial features. Additionally, we perform CLIP classification and manual screening to verify the correctness of both the attribute class and occupation in the generated samples.

\noindent \textbf{Implementation Details.}
Our main experiments are conducted on the SD1.5, while the generalizability of our approach is also demonstrated on more advanced models, SD2.1~\citep{Rombach2022stable} and SDXL~\citep{podell2023sdxl}. By default, the inclusive token $v_i$ is initialized to the embedding of the token ``individual'', which we identify as a natural inclusive token (further discussion on natural inclusive tokens is provided in Appendix Sec.~\ref{sec:nit-init}).
The adaptive inclusive token is set to a length of 1, resulting in $v_i^c \in \mathbb{R}^{1 \times 768}$, as $768$ is the token embedding dimension in the SD1.5 text model. 
Training is conducted for \major{1 epoch with 15 times repeat, using a total batch size of 4.}
This setup results in a total training time of approximately 1 hour on a single NVIDIA A100 GPU. The model is optimized using the AdamW optimizer~\citep{loshchilov2018decoupled} with a base learning rate of $5\times 10^{-4}$.
For training prompts, we follow the approach of Textual Inversion~\citep{gal2022textual}, utilizing ImageNet-style~\citep{imagenet} templates to describe objects while incorporating adjectives to specify the attribute of interest. The complete list of template prompts is provided in Appendix Sec.~\ref{sec:prompttemp}.
The adaptive mapping network follows a transformer architecture, consisting of six attention heads and four transformer blocks.

\noindent \textbf{Evaluation Protocols.}
To assess the effectiveness of our method in mitigating stereotypical biases in T2I generation, we construct a test set comprising 24 unseen occupations that are excluded from the training set.
For each biased concept, we analyze 100 generated images with valid detected faces to measure the distribution of different attribute classes. All evaluation images are generated using the test prompt: ``a photo of a $\mathord{<}i\mathord{>}$ \{occupation\}'' with 25 denoising steps. 
Additionally, to better balance the six racial categories, the racial inclusive token $\mathord{<}i_r\mathord{>}$ is introduced after 10 sampling steps on the base prompt $T(c)$.
To quantify fairness, attribute classifiers are employed to detect and compute the distribution of different attribute classes in the generated images. Following prior works~\citep{zhang2023itigen, gandikota2023unified, kim2023destereotyping}, we adopt the CLIP~\citep{radford2021clip} zero-shot classifier to classify sensitive attributes, using the prompt: ``a photo of a [attribute] person''.

\begin{table*}[tb]
  \caption{Comparisons with baseline methods on fairness, quality, and text alignment of generative results across three bias attributes \textcolor{black}{(See evaluation metrics details in Sec.~\ref{sec:setup})}. Abbreviations are used for simplicity: \textbf{SD1.5}: \citet{Rombach2022stable}. \textbf{ITI-GEN}: \citet{zhang2023itigen}. \textbf{TIME}: \citet{orgad2023time}. \textbf{FD}: \citet{friedrich2023fairdiffusion}. \textbf{EI}: \citet{bansal2022howwell}. \textbf{FM}: \citet{li2023fairmapping}. \textbf{rTI}: \citet{gal2022textual}. $^*$ indicates editing-based methods that require careful tuning of editing strengths to achieve reasonable results. \textcolor{black}{The best and second best results among methods requiring no attribute \minor{class} specification \minor{during inference} or prior knowledge on the bias distribution are marked by \colorbox{cyan!25}{blue} and \colorbox{orange!25}{orange}, respectively.}
  }
  \centering
  \renewcommand\arraystretch{1.2}
  \normalsize
  \setlength{\tabcolsep}{2.16mm}
	{
  \begin{tabular}{l|ccc|ccc|ccc}
        \hline
         & \multicolumn{3}{c|}{\textbf{Gender}} & \multicolumn{3}{c|}{\textbf{Race}} & \multicolumn{3}{c}{\textbf{Age}}\\
        \cline{2-10}
        \multirow{-2.0}{*}{\textbf{Methods}} & $\mathcal{D_{KL} \downarrow}$ & FID$ \downarrow$ & CLIP$ \uparrow$ &  $\mathcal{D_{KL} \downarrow}$ & FID$ \downarrow$ & CLIP$ \uparrow$ &  $\mathcal{D_{KL} \downarrow}$ & FID$ \downarrow$ & CLIP$ \uparrow$ \\
        \hline
        SD1.5 & 0.3584 & 281.12 & 0.2823 & 0.5973 & 281.12 & 0.2823 & 0.2319 & 281.12 & 0.2823 \\
        \hline
        \multicolumn{10}{l}{w/ attribute \minor{class} specification \minor{during inference} or prior knowledge on the bias distribution} \\
        \hline
        ITI-GEN & 0.0078 & 278.21 & 0.2753 & 0.3699 & 247.05 & 0.2679 & 0.1560 & 243.09 & 0.2648 \\
        TIME$^*$ & 0.2908 &	277.79 & 0.2733 & 0.5463 &	270.03 & 0.2663 & 0.2285 & 271.09 & 0.2738 \\
        FD$^*$ & 0.2420 &	278.10 & 0.2718 & 0.4987  & 277.64 & 0.2738 & 0.2246 & 280.33 & 0.2740 \\
        \hline
        \multicolumn{10}{l}{w/o attribute \minor{class} specification \minor{during inference} or prior knowledge on the bias distribution} \\
        \hline
        EI & 0.1666	& 283.52 & \cellcolor{orange!25} 0.2758 & 0.6033 & 281.11 & 0.2745 & \cellcolor{orange!25}0.2258	& 289.82 & 0.2773 \\
        FM & \cellcolor{cyan!25}0.1174	& \cellcolor{cyan!25}222.82 & 0.2341 & \cellcolor{orange!25}0.3722 & \cellcolor{cyan!25}220.37 & 0.2391 & 0.3823 & \cellcolor{cyan!25}255.72 & 0.2402 \\
        rTI & \major{0.3124} & \cellcolor{orange!25}\major{261.98} & \major{0.2751} & \major{0.8230} & \cellcolor{orange!25}\major{274.66} & \cellcolor{orange!25}\major{0.2791} & \major{0.3711} & \major{287.12} & \cellcolor{cyan!25}\major{0.2795} \\
        Ours & \cellcolor{orange!25}\major{0.1290} & \major{268.55} & \cellcolor{cyan!25}\major{0.2804} & \cellcolor{cyan!25}\major{0.3488} & \major{278.77} & \cellcolor{cyan!25}\major{0.2813} & \cellcolor{cyan!25}\major{0.2135} & \cellcolor{orange!25}\major{274.12} & \cellcolor{orange!25}\major{0.2789} \\
        \hline
    \end{tabular}
    }
    \label{tab:baseline}
\end{table*}

\begin{table}[!t]
  \caption{Performance on SD 2.1 and SDXL.}
  \centering
  \renewcommand\arraystretch{1.2}
  \setlength{\tabcolsep}{.9mm}
 {
  \begin{tabular}{c | ccc}
        \hline
        \textbf{Models} & \textbf{Gender} $\mathcal{D_{KL}}$ & \textbf{Race} $\mathcal{D_{KL}}$ & \textbf{Age} $\mathcal{D_{KL}}$ \\
        \hline
        SD2.1  & 0.4166 & 0.5819 & 0.2062 \\
        SD2.1 w/ ours & \major{\textbf{0.2552}} & \major{\textbf{0.3727}} & \major{\textbf{0.1748}} \\
        \hline
        SDXL & 0.4919 & 0.6796 & 0.1965 \\
        SDXL w/ ours & \major{\textbf{0.2009}} & \major{\textbf{0.4441}} & \major{\textbf{0.1223}} \\
        \hline
    \end{tabular}
    }
    \label{tab:dm21}
\end{table}

\noindent \textbf{Evaluation Metrics.} 
We employ three metrics to evaluate the effectiveness of our approach. (1) \textbf{Fairness Assessment}: following \citet{zhang2023itigen}, we measure the fairness of the debiased attribute distribution using KL divergence ($\mathcal{D_{KL}}$)$\downarrow$ against an even distribution. (2) \textbf{Image Quality}: We quantify the perceptual quality of generated images using the Fr\'{e}chet Inception Distance (FID)$\downarrow$~\citep{heusel2017fid}. (3) \textbf{Text-Image Alignment}: To evaluate the extent of concept drifting, we compute the CLIP-Score (CLIP)$\uparrow$~\citep{radford2021clip}, which measures the alignment between generated images and their corresponding prompts.

\begin{figure*}[t]
    \centering
    \includegraphics[width=\textwidth]{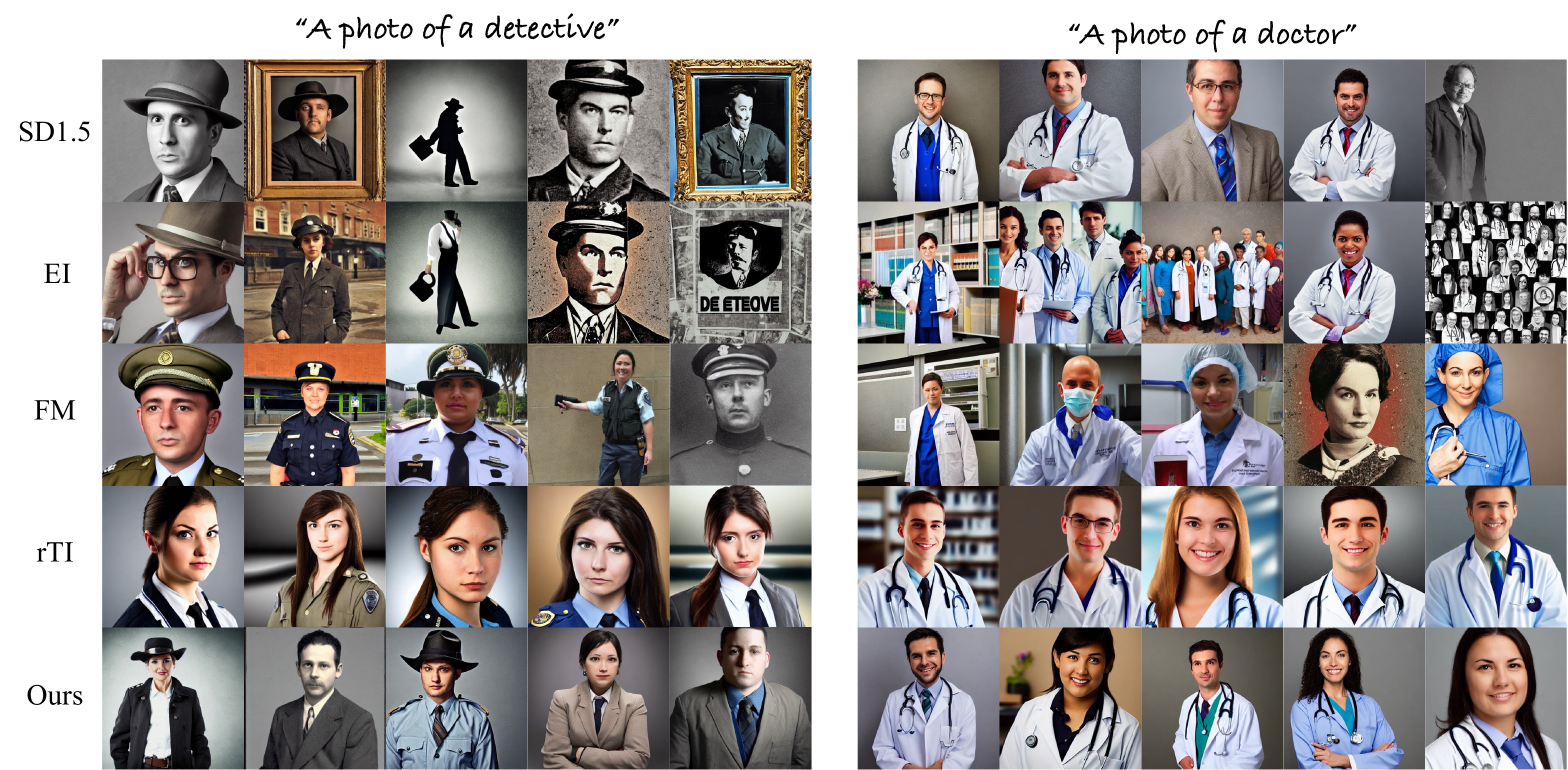}
        \caption{Qualitative evaluation on gender bias mitigation of \textbf{stereotypically male-dominated occupations}. All images are generated with the same random seed. The captions above indicate the base prompt $T(c)$.}
    \label{fig:gender-male}

    \centering
    \includegraphics[width=\textwidth]{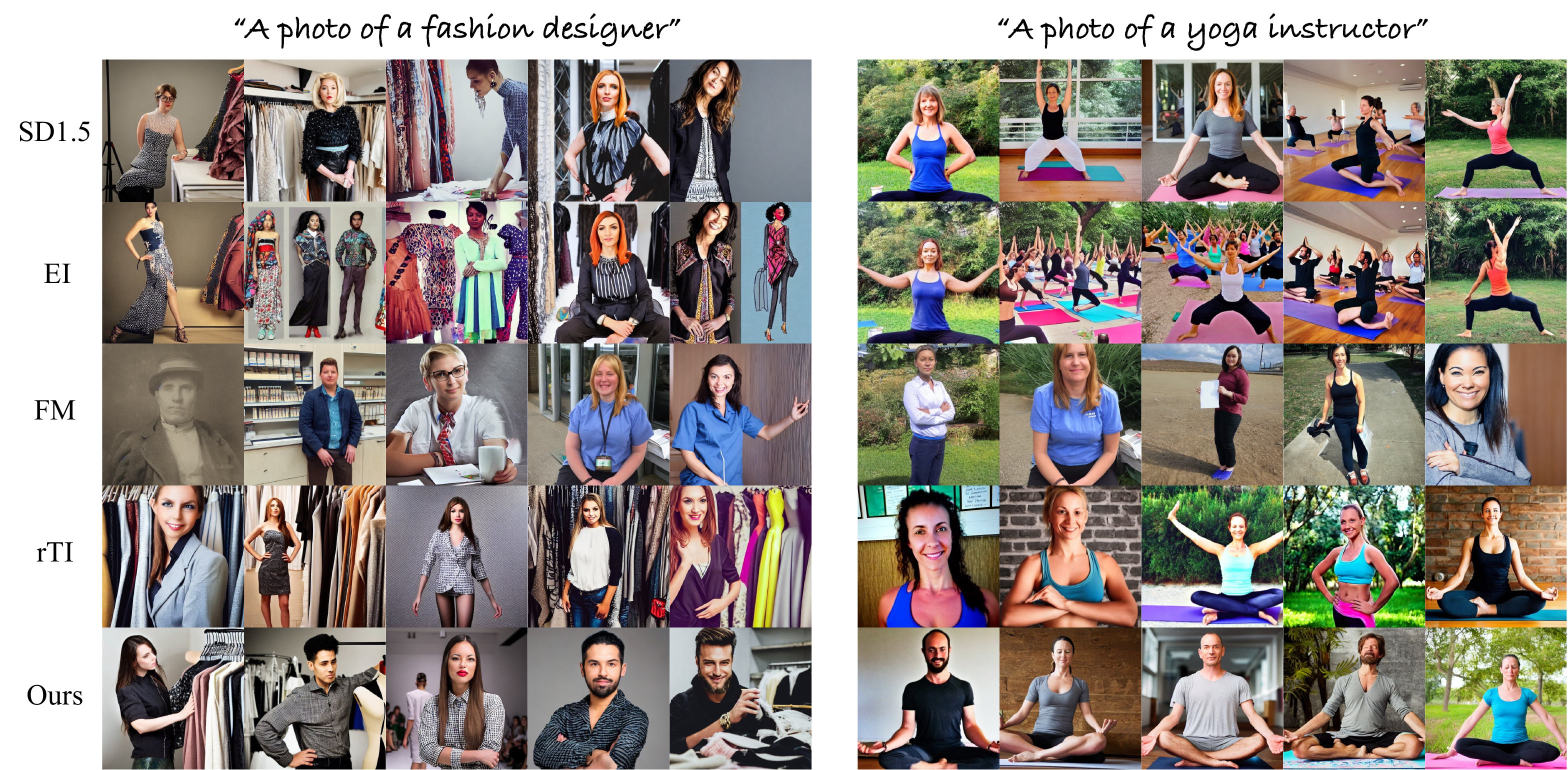}
        \caption{Qualitative evaluation on gender bias mitigation of \textbf{stereotypically female-dominated occupations}. All images are generated with the same random seed. The captions above indicate the base prompt $T(c)$.}
    \label{fig:gender-female}
\end{figure*}

\begin{figure*}
    \centering
    \includegraphics[width=\textwidth]{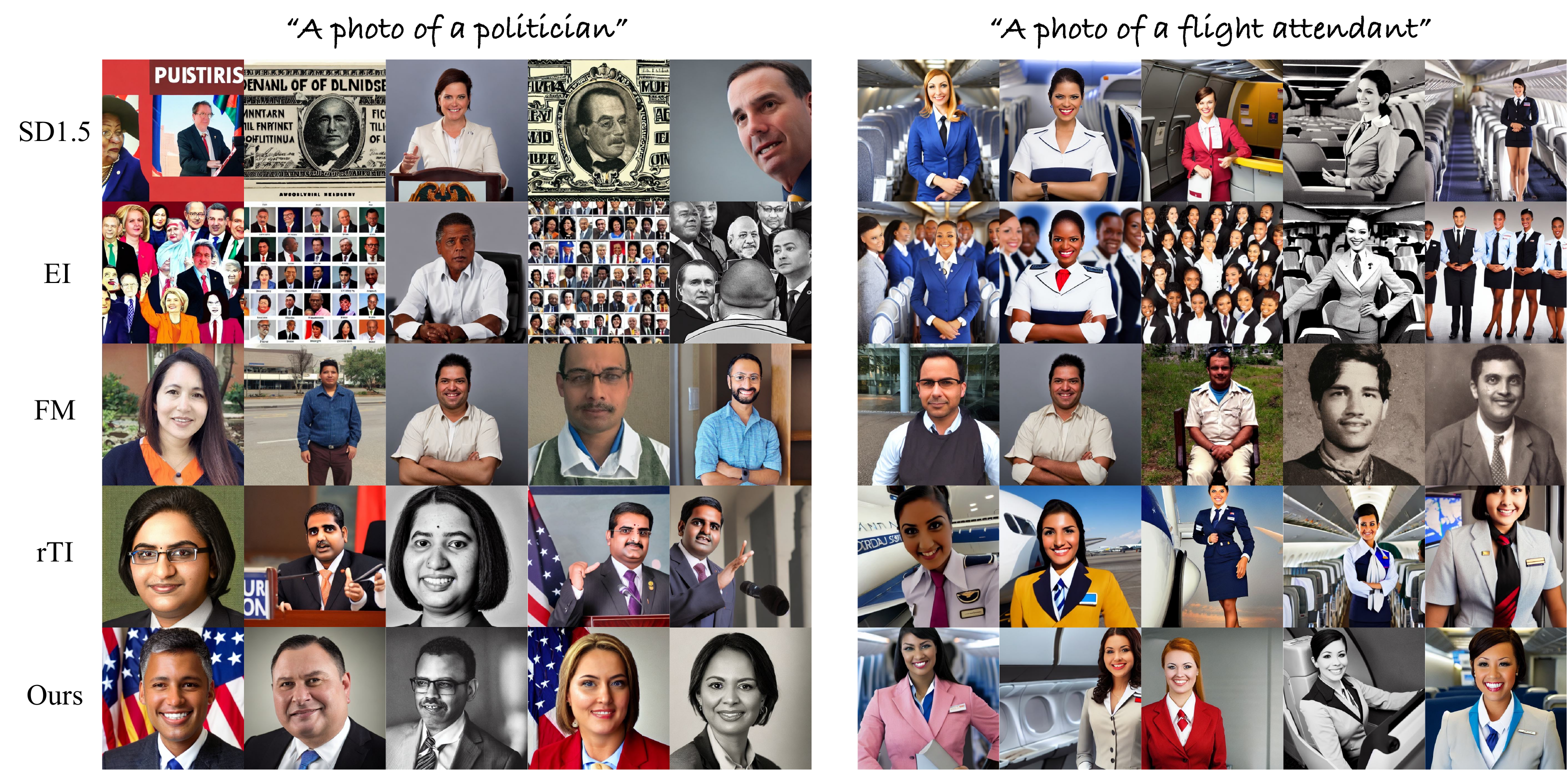}
        \caption{Qualitative evaluation on race biases mitigation. All images are generated with the same random seed. The captions above indicate the base prompt $T(c)$.}
    \label{fig:race}

    \centering
    \includegraphics[width=\textwidth]{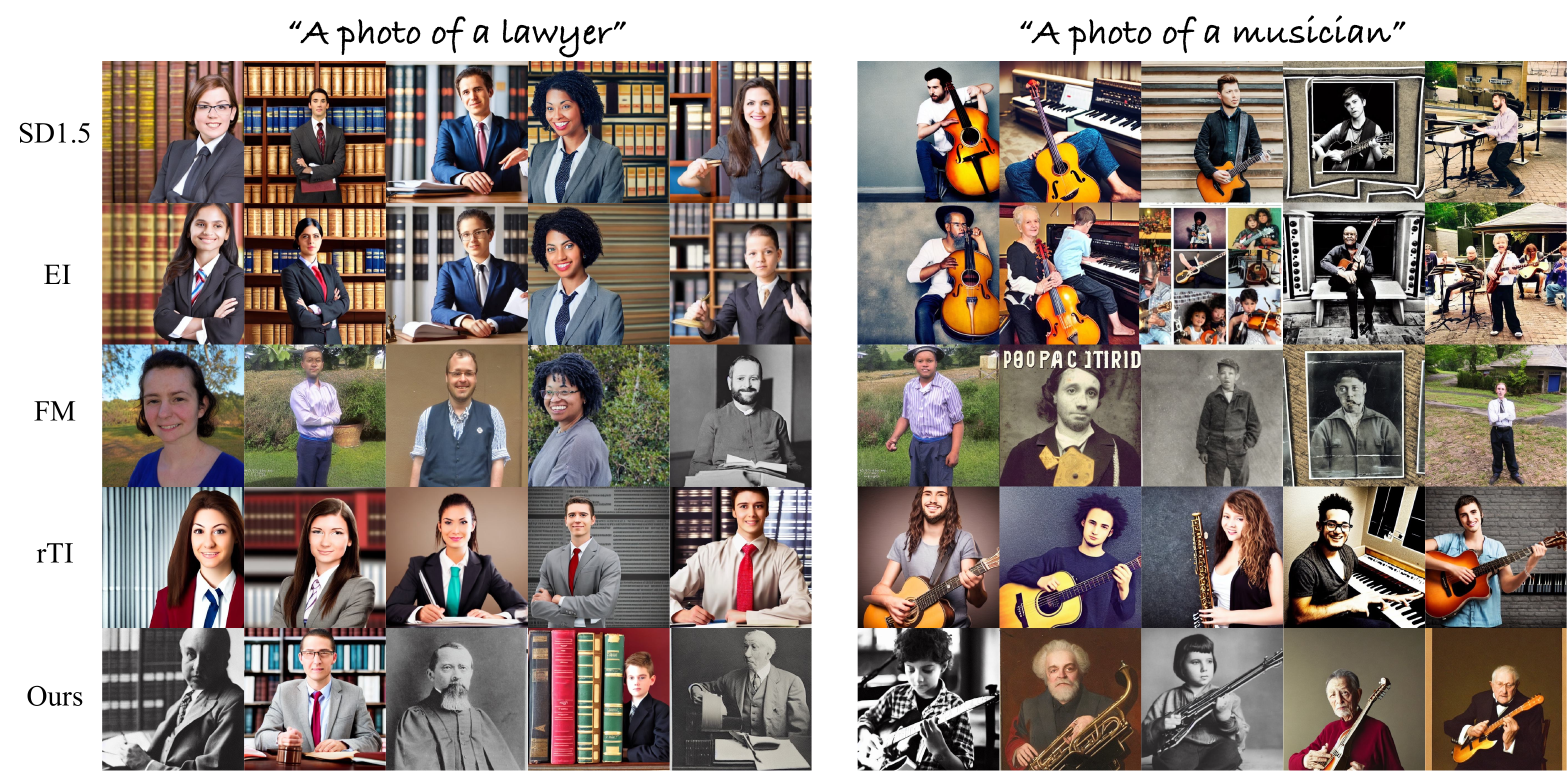}
        \caption{Qualitative evaluation on age biases mitigation. All images are generated with the same random seed. The captions above indicate the base prompt $T(c)$.}
    \label{fig:age}
\end{figure*}

\noindent \textbf{Baselines}
\label{sec:baseline}
We compare our method with several baseline approaches, which are briefly described as follows: (1) \textbf{Stable Diffusion} (SD1.5)~\citep{Rombach2022stable}: The original SD1.5 pipeline is used with the base prompt $T(c)=$ ``A photo of a \{occupation\}''. (2) \textbf{ITI-GEN}~\citep{zhang2023itigen}: ITI-GEN is trained separately for each attribute using the additional datasets provided in their work. During generation, it loops over all attribute classes and generates an equal number of images per class to ensure an even distribution. When the number of classes is not divisible by 100, a slightly larger number of images is generated to maintain balance. Since race was not explicitly studied in ITI-GEN, we perform race classification on their inclusive skin tone generation while acknowledging its limitations. (3) \textbf{TIME}~\citep{orgad2023time}: Since TIME cannot generalize to unseen concepts, we train a separate model for each test occupation. The editing strength parameter $\lambda$ is tuned from $[10, 1, 0.1]$ to select the best-performing value for each occupation. (4) \textbf{Fair Diffusion}~\citep{friedrich2023fairdiffusion}: We follow the default editing parameters provided by the authors. For gender and age (binary attributes), we edit the model toward the non-stereotypical class.
For race (six classes), we loop over all six classes as editing directions to obtain a comprehensive evaluation. (5) \textbf{Ethical Intervention}~\citep{bansal2022howwell}: The best-performing intervention ``...if all individual can be a \{occupation\} irrespective of their \{attribute\}'', as reported in their work, is added to the base prompt. (6) \textbf{Fair Mapping}~\citep{li2023fairmapping}: The model is trained separately for each attribute following the implementation details provided until the training loss converges. (7) Revised \textbf{Textual Inversion}~\citep{gal2022textual}: This baseline follows the revised version discussed in Sec.~\ref{sec:preliminary}.

\begin{figure*}[!t]
    \centering
    \includegraphics[width=0.95\textwidth]{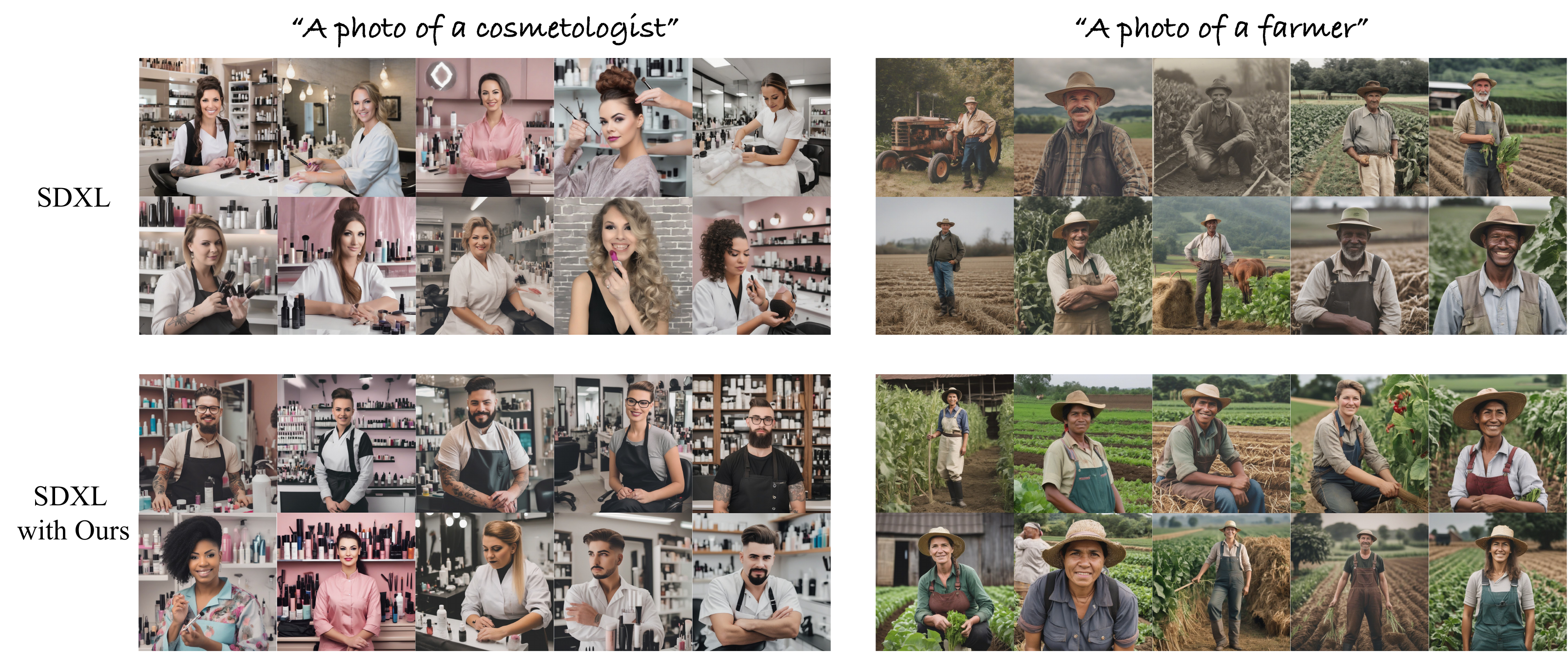}
    \caption{Results of gender bias mitigation on SDXL. All images are generated with the same random seed.}
    \label{fig:sdxl-gender}
    \centering
    \includegraphics[width=0.95\textwidth]{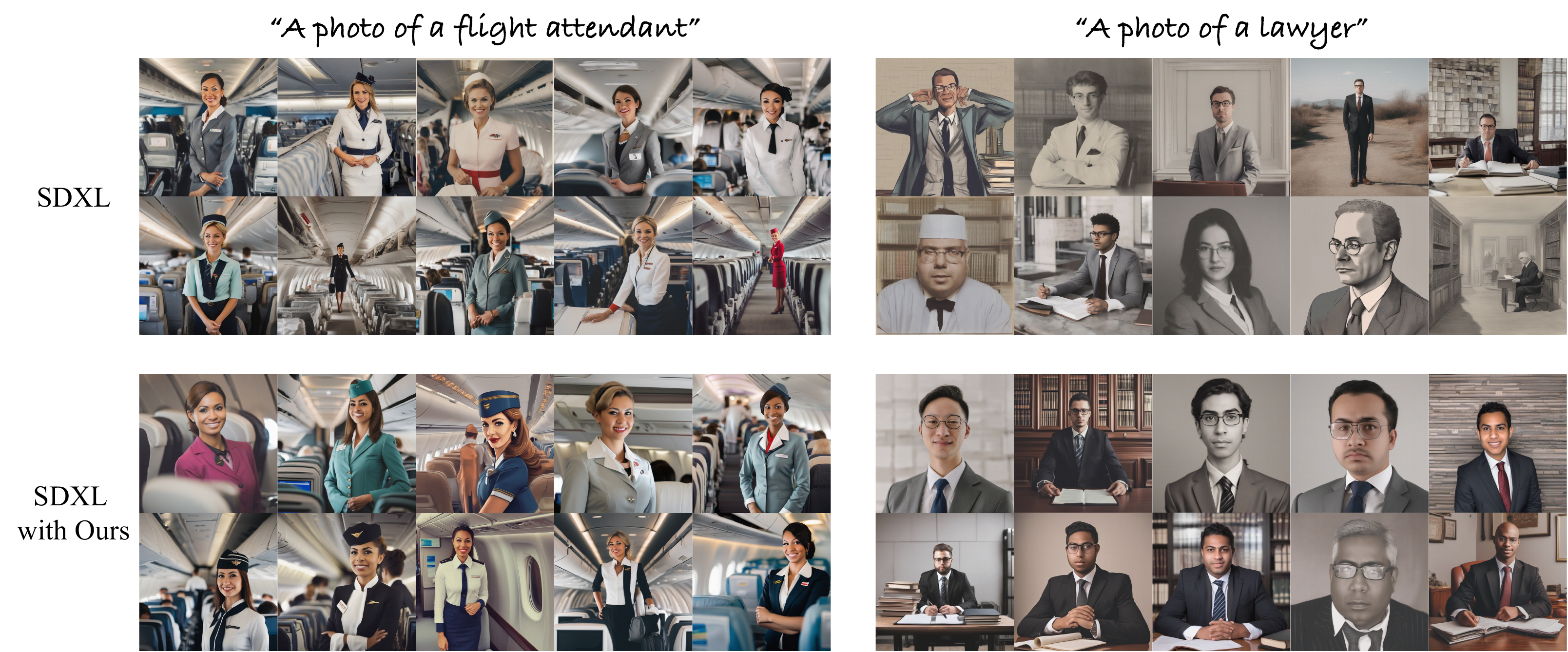}
    \caption{Results of race bias mitigation on SDXL. All images are generated with the same random seed.}
    \label{fig:sdxl-race}
    \centering
    \includegraphics[width=0.95\textwidth]{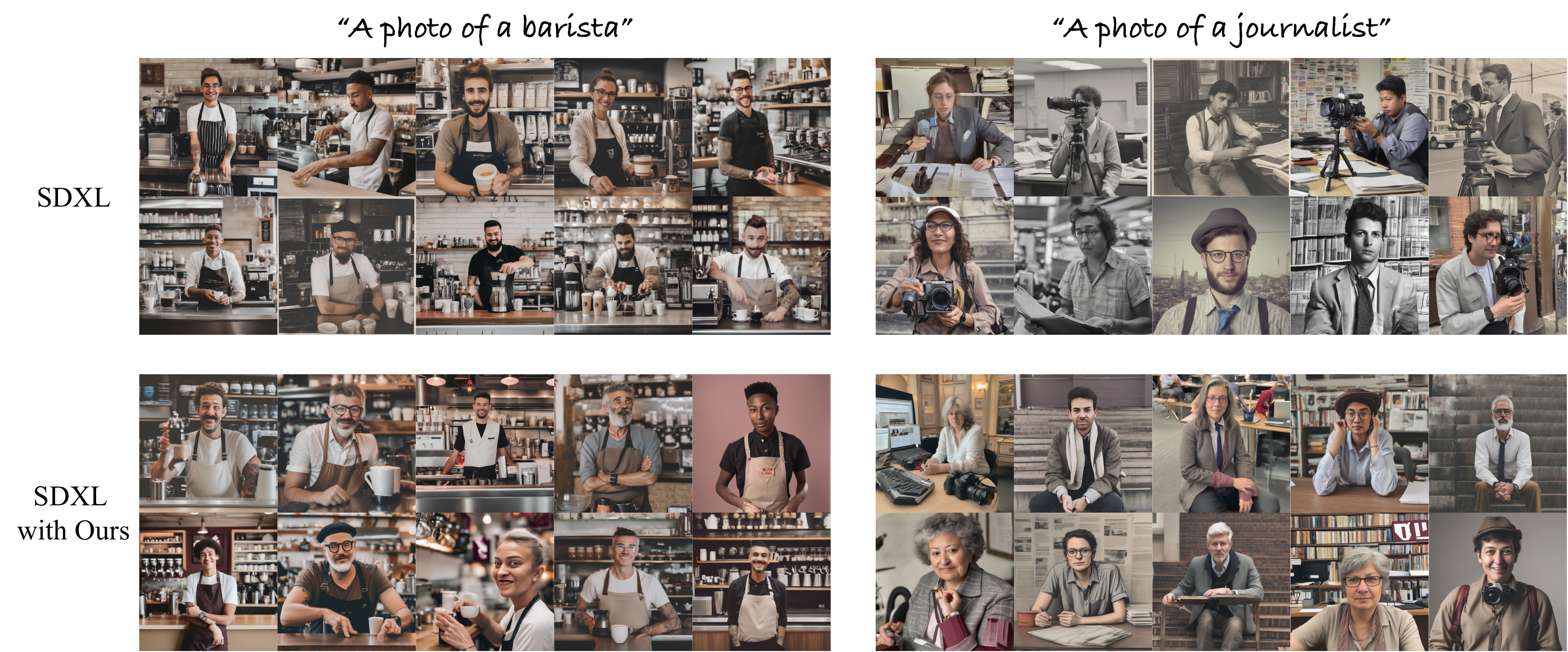}
    \caption{Results of age bias mitigation on SDXL. All images are generated with the same random seed.}
    \label{fig:sdxl-age}
\end{figure*}

\subsection{Experimental Results}

\minor{We organize our experimental results into five parts, focusing on overall performance, key components, generalization behavior, design choices, and the impact of training data.}

\subsubsection{\minor{Main Performance}}
Table~\ref{tab:baseline} presents a quantitative comparison of our proposed method against baseline approaches for mitigating single bias. The methods are categorized into two groups: those that rely on attribute \minor{class} specification or editing direction guidance, and those that operate without additional attribute information. 
As shown in the table, methods with explicit attribute \minor{class} specification generally achieve strong inclusiveness. However, as discussed in Sec.~\ref{sec:introduction}, achieving fairness without specifying attribute classes or editing directions is a more meaningful and challenging task. Our approach excels in mitigating race and age biases without requiring additional guidance and achieves comparable performance to the best-performing method in gender bias mitigation based on numerical metrics.
Qualitative comparisons of approaches that do not rely on additional information are presented in Figs.~\ref{fig:gender-male}, \ref{fig:gender-female}, \ref{fig:race}, and \ref{fig:age}. Our method effectively enhances inclusiveness while preserving the semantic meaning of a given concept.

Despite the lower distribution discrepancy reported for Fair Mapping (FM)~\citep{li2023fairmapping} in gender bias, its CLIP-Score experiences a significant drop, indicating a loss of visual semantics related to the target occupation in the generated images. This semantic drift is clearly visible in the qualitative results (Figs.~\ref{fig:gender-male}, \ref{fig:gender-female}, \ref{fig:race}, \ref{fig:age}). Similarly, Ethical Intervention (EI)~\citep{bansal2022howwell}, which achieves comparable performance to our method in age bias mitigation, frequently generates multiple individuals in a single image to satisfy its inclusive constraint. However, this behavior contradicts the given prompt, which explicitly requests ``\textbf{a} \{occupation\}'', implying the presence of a single individual.

\begin{table}[tb]
  \caption{Ablation studies of the key components.}
  \centering
  \renewcommand\arraystretch{1.2}
   \setlength{\tabcolsep}{1.1mm}
  {
  \begin{tabular}{cc | ccc}
        \hline
         $L_{anchor}$ & $\mathcal{F}_{am}$ & \textbf{Gender} $\mathcal{D_{KL}}$ & \textbf{Race} $\mathcal{D_{KL}}$ & \textbf{Age} $\mathcal{D_{KL} }$ \\
         \hline
        & & \major{0.3124} & \major{0.8230} & \major{0.3711} \\
        \checkmark & & \major{0.3121} & \major{\textbf{0.3461}} & \major{0.3225} \\
        & \checkmark & \major{0.3063} & \major{0.6172} & \major{0.3412} \\
        \checkmark & \checkmark & \major{\textbf{0.1290}} & \major{0.3488} & \major{\textbf{0.2135}} \\
        \hline
    \end{tabular}
    }
    \label{tab:comp_ablation}
    \hspace{-10pt}
\end{table}

\begin{table*}[tb]
  \caption{Performance of combining adaptive inclusive tokens to mitigate multi-biases.} 
  \centering
  \renewcommand\arraystretch{1.2}
   \setlength{\tabcolsep}{1.5mm}
  {
  \begin{tabular}{ccc | ccc | cc}
        \hline
         $\mathord{<}i_g\mathord{>}$ & $\mathord{<}i_r\mathord{>}$ & $\mathord{<}i_a\mathord{>}$ & \textbf{Gender} $\mathcal{D_{KL} \downarrow}$ & \textbf{Race} $\mathcal{D_{KL} \downarrow}$ & \textbf{Age} $\mathcal{D_{KL} \downarrow}$ & \textbf{FID}$ \downarrow$ & \textbf{CLIP}$ \uparrow$ \\
         \hline
          &  &  & 0.3584 & 0.5973 & \textbf{0.2319} & 281.12 & \textbf{0.2823} \\
         \checkmark & \checkmark & \checkmark & \major{\textbf{0.1812}} & \major{\textbf{0.4626}} & \major{0.2790} & \major{\textbf{262.11}} & \major{0.2786} \\
        \hline
    \end{tabular}
    }
    \label{tab:multi}
    \vspace{6pt}
\end{table*}

Furthermore, our method is model-agnostic and generalizes to any T2I model that supports a text encoder capable of handling newly learned embeddings. To validate this, we evaluate our approach on SD2.1\footnote{\texttt{stabilityai/stable-diffusion-2-1}}~\citep{Rombach2022stable} and SDXL\footnote{\texttt{stabilityai/stable-diffusion-xl-base-1.0}}~\citep{podell2023sdxl}.
Table~\ref{tab:dm21} demonstrates the effectiveness of our method on SD2.1, confirming its generalizability.
For SDXL, training data is generated following the data preparation procedure in Sec.~\ref{sec:train data} using the SDXL model itself. The adaptive mapping module is integrated into the first text model of SDXL, while the second text model remains unchanged. The results are presented in Tab.~\ref{tab:dm21}, with qualitative results shown in Figs.~\ref{fig:sdxl-gender}, \ref{fig:sdxl-race}, and \ref{fig:sdxl-age}. Both quantitative and qualitative results demonstrate that our adaptive inclusive token method effectively generalizes to more advanced SD models, achieving more inclusive T2I outcomes.

\begin{table*}[tb]
\caption{Performance of adaptive inclusive tokens in complex scenes. Reported metrics are on gender bias. Additional prompts are added to ``A photo of a $\mathord{<}i_g\mathord{>}$ \{occupation\}''.}
  \centering
  \renewcommand\arraystretch{1.2}
    \setlength{\tabcolsep}{2.8mm}{
  \begin{tabular}{c | ccc | ccc | ccc}
        \hline
         \textbf{Metrics} & $\mathcal{D_{KL} \downarrow}$ & FID$ \downarrow$ & CLIP$ \uparrow$ & $\mathcal{D_{KL} \downarrow}$ & FID$ \downarrow$ & CLIP$ \uparrow$& $\mathcal{D_{KL} \downarrow}$ & FID$ \downarrow$ & CLIP$ \uparrow$\\
         \hline
         Prompts & \multicolumn{3}{c|}{+ ``drinking coffee.''} & \multicolumn{3}{c|}{+ ``reading a book.''} & \multicolumn{3}{c}{+ ``listening to music.''}\\
         \hline
         SD1.5 & 0.3567 & 266.75 & \textbf{0.3116} & 0.3992 & 326.80 & \textbf{0.3045} & 0.4246 & \textbf{274.67} & \textbf{0.3033} \\
         Ours & \major{\textbf{0.1679}} & \major{\textbf{259.24}} & \major{0.2453}  & \major{\textbf{0.2643}} & \major{\textbf{310.90}} & \major{0.2468} & \major{\textbf{0.2263}} & \major{276.93} & \major{0.2473} \\
        \hline
    \end{tabular}
    }
    \label{tab:complex}
\end{table*} 

\subsubsection{\minor{Key Component Analysis}}
We conduct ablation studies to evaluate the effectiveness of our proposed components $L_{anchor}$ and $\mathcal{F}_{am}$. The quantitative results are presented in Tab.~\ref{tab:comp_ablation}.
As shown in the table, the combination of both proposed components generally achieves the best performance, except in the case of race bias. We hypothesize that this is due to the homogeneous bias distribution in many occupations, where the majority of generated figures are biased toward White individuals. This dominance likely reduces the effectiveness of the adaptive mapping network, as it relies on variation across attribute classes to demonstrate effective bias corrections.

\subsubsection{\minor{Generalization Analysis}}

\textbf{Multiple Biases.}
To evaluate the ability of our adaptive inclusive tokens to mitigate multiple stereotypical biases simultaneously, we concatenate previously single-attribute-trained inclusive tokens during inference using the prompt: ``A photo of a $\mathord{<}i_g\mathord{>}$ $\mathord{<}i_r\mathord{>}$ $\mathord{<}i_a\mathord{>}$ \{occupation\}''. As shown in Tab.~\ref{tab:multi}, the adaptive inclusive tokens are effectively combined to promote inclusive generation across gender and race attributes, demonstrating their scalability and flexibility in bias mitigation. \major{It is important to note that the combination of all three adaptive tokens results in a marginal increase in the age bias (from 0.2319 to 0.2790). This outcome is anticipated, as the token training data for $\mathord{<}i_g\mathord{>}$ and $\mathord{<}i_r\mathord{>}$ were not subject to strict control over age distribution. Specifically, the inclusion of $\mathord{<}i_g\mathord{>}$ and $\mathord{<}i_r\mathord{>}$ introduces age-related spurious correlations learned from their respective training sets. Looking ahead, if training data balanced across all three attributes (gender, race, and age) can be collected, we can train a single, unified token, $\mathord{<}i_{all}\mathord{>}$, to simultaneously debias across all three attributes.}

\noindent \textbf{Complex Scenes.}
To evaluate the generalizability of our adaptive inclusive tokens to unseen and more complex prompts, we introduce three additional prompts describing individuals engaged in various activities. This evaluation is conducted using the $\mathord{<}i_g\mathord{>}$ token only as a preliminary demonstration. 
The results presented in Tab.~\ref{tab:complex}, indicate that our adaptive inclusive token effectively extends to more complex scenarios, achieving comparable text-image alignment while maintaining or even improving image quality.

\begin{table}[!t]
    \centering
    \caption{\major{Ablation study on adaptive mapping network architecture.}}
    \begin{tabular}{c|ccc}
        \hline
        Network & Gender $\mathcal{D_{KL} \downarrow}$ & FID$ \downarrow$ & CLIP$ \uparrow$ \\
        \hline
        MLP & 0.4059 & \textbf{267.17} & 0.2788 \\
        Transformer & \textbf{0.1290} & 268.55 & \textbf{0.2804} \\
        \hline
    \end{tabular}
    \label{tab:mlp}
\end{table}

\begin{table*}[tb]\
  \caption{Ablation study on the complexity of the adaptive mapping network.}
  \centering
  \renewcommand\arraystretch{1.2}
    \setlength{\tabcolsep}{1.6mm}{
  \begin{tabular}{cc | ccc}
        \hline
         \textbf{Transformer Blocks} & \textbf{Attention Heads} & \textbf{Gender} $\mathcal{D_{KL} \downarrow}$ & \textbf{Race} $\mathcal{D_{KL} \downarrow}$ & \textbf{Age} $\mathcal{D_{KL} \downarrow}$ \\
         \hline
         \multicolumn{2}{c|}{SD1.5 Original Pipeline} & 0.3584 & 0.5973 & 0.2319 \\
         \hline
        4 & 6 & \major{\textbf{0.1290}} & \major{0.3488} & \major{0.2135} \\
        2 & 6 & \major{0.1630} & \major{\textbf{0.3348}} & \major{\textbf{0.1887}} \\
        4 & 8 & \major{0.1337} & \major{0.3468} & \major{0.2192} \\
        2 & 8 & \major{0.1717} & \major{0.3391} & \major{0.1993} \\
        \hline
    \end{tabular}
    }
    \label{tab:complexity}
\end{table*}

\begin{table*}[tb]
  \caption{Ablation study on the placement of the adaptive mapping module. Reported metrics are on gender bias.}
  \centering
  \renewcommand\arraystretch{1.2}
    \setlength{\tabcolsep}{2.7mm}{
  \begin{tabular}{c | ccc}
        \hline
         \textbf{Placement of Adaptive Mapping} & $\mathcal{D_{KL} \downarrow}$ & FID $\downarrow$ & CLIP $\uparrow$ \\
         \hline
        Before Text Transformer & \major{\textbf{0.1290}} & \major{\textbf{268.55}} & \major{\textbf{0.2804}} \\
        After Text Transformer & \major{0.3388} & \major{271.67} & \major{0.2801} \\
        \hline
    \end{tabular}
    }
    \label{tab:place}
\end{table*}

\subsubsection{\minor{Design Choice Analysis}}

\noindent \textbf{Adaptive Mapping Network Architecture.}
We further investigate the choice of network architecture for the adaptive mapping module by comparing a Transformer with a multi-layer perceptron (MLP), as shown in Tab.~\ref{tab:mlp}. To ensure a fair comparison, we control the number of parameters so that both networks are of comparable size. Despite having similar capacity, the Transformer achieves better bias mitigation while maintaining competitive fidelity and alignment scores. The key advantage lies in its ability to naturally handle variable-length input sequences and capture semantic dependencies across tokens through self-attention. In contrast, the MLP requires fixed-size input vectors. In our ablation, this limitation was addressed by padding sequences to a fixed length. However, such padding reduces flexibility and can cause the model to overfit to non-semantic artifacts, while overlooking meaningful relationships between tokens. These results highlight that the Transformer architecture is inherently more suited for adapting inclusive tokens, as it provides both the flexibility to process diverse textual contexts and the capacity to capture fine-grained dependencies essential for effective bias mitigation.

\noindent \textbf{Adaptive Mapping Network Complexity.}
We employ a transformer architecture for the adaptive mapping network and examine the effect of varying its complexity, as shown in Tab.~\ref{tab:complexity}. The results indicate that our adaptive inclusive token effectively reduces stereotypical bias in T2I outputs regardless of the network complexity. 

\major{The optimal complexity, however, varies by attribute. Gender bias mitigation is most effective with the deeper network configuration, suggesting a need for increased capacity to learn the necessary representations. Conversely, the best performance for both race and age biases is achieved using the shallower 2-block setup, indicating that excessive complexity is unnecessary for these attributes. Overall, the 4 blocks, 6 heads configuration offers the most balanced compromise, achieving optimal gender debiasing while retaining highly competitive race and age results. Thus, we adopt this setting as our primary one.}

\minor{This variation in optimal complexity can be attributed to differences in the distributions of different bias types. For gender bias, although the attribute is binary, it exhibits strong semantic entanglement with occupations: certain occupations are stereotypically associated with male while others with female representations. As a result, the adaptive mapping network must learn a non-linear mapping that detects the direction of bias encoded in the concept embedding $v_c$ and generates a correction token $v_i^c$ that counteracts it. This ``direction switching'' requires greater representational capacity, which is effectively supported by deeper networks (\eg, 4 transformer blocks). In contrast, race and age bias are often dominated by a single majority class, leading to a simpler adaptation task. For these attributes, shallower networks are sufficient to model the bias correction function.}

\begin{figure}[t!]
    \centering
    \includegraphics[width=0.98\linewidth]{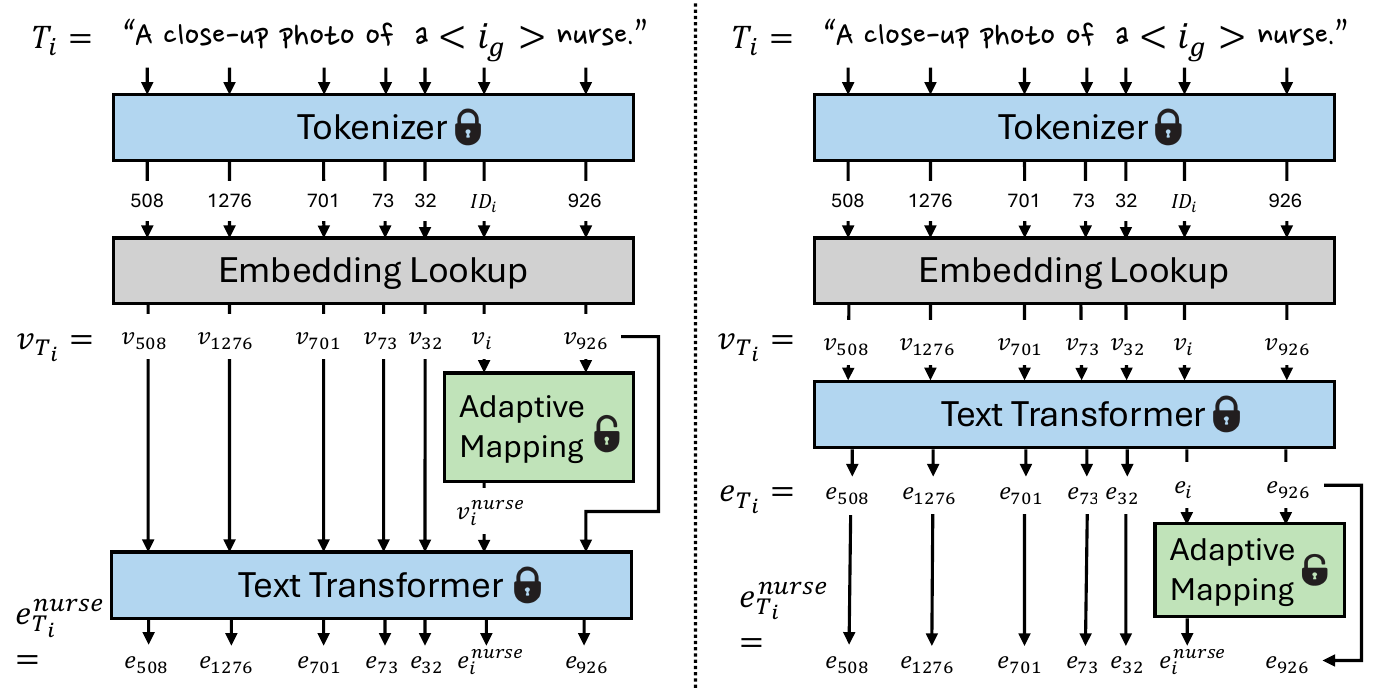}
    \caption{Various placements of the adaptive mapping module. \textbf{Left}: before text transformer (inclusive embedding in token embeddings). \textbf{Right}: after text transformer (inclusive embedding in text embeddings).}
    \label{fig:placement}
\end{figure}

\noindent \textbf{Placement of Adaptive Mapping Module.}
The placement of the adaptive inclusive embedding within the text model can significantly impact its effectiveness in bias mitigation. To examine this effect, we conduct experiments where the adaptive mapping module is positioned before and after the text transformer, as illustrated in Fig.~\ref{fig:placement}. The quantitative results, presented in Tab.~\ref{tab:place}, indicate that placing the adaptive mapping module after the text transformer reduces the influence of the inclusive embedding on the final outputs, thereby diminishing its effectiveness in bias mitigation.

To further investigate this phenomenon, we conduct two additional experiments analyzing the effect of embeddings before and after the text transformer. Here, we denote: the token embeddings before the text transformer as $v_{T_i}$, and the text embeddings after the text transformer as $e_{T_i}$. 
First, for a neutral occupation-related prompt ``A photo of a \{occupation\}'' where no gender class is specified, we find that most generated images still reflect the stereotypical gender associated with the occupation—even when the non-stereotypical gender embedding is injected after the text transformer. This suggests that gender bias is already embedded throughout the entire text embedding via the text transformer, making late-stage modifications less effective.
To further validate this, we modify the prompt by explicitly specifying a non-stereotypical gender class, \eg, ``A photo of a male nurse''. We then replace the text embedding of ``male'' with that of ``female'', extracted from a separate prompt: ``A photo of a female nurse'' after the text transformer. Surprisingly, the generated images still predominantly depict male nurses, despite the late-stage modification. This result indicates that attribute indicators have a greater impact when introduced at the token embedding level, as they propagate more effectively throughout the entire text embedding via the transformer architecture.
These findings align with our quantitative observations, confirming that the adaptive inclusive token influences the final output attributes more effectively when the adaptive mapping module is placed before the text transformer.

\noindent \textbf{Number of Inclusive Tokens.}
We explore the impact of varying the number of learned tokens in carrying the inclusiveness requirement. Tab.~\ref{tab:no} demonstrates that a single adaptive token is sufficient to achieve a more inclusive attribute distribution in the generated outputs, reinforcing the effectiveness of our approach.
\minor{This observation is closely related to the nature of the bias attributes considered in this work. Sensitive attributes such as gender, race, and age of a person are primarily expressed through subtle visual cues (\eg, hair, beard, wrinkle, etc.), rather than through large-scale semantic changes. As a result, effective bias mitigation can often be achieved through a relatively small and targeted adjustment in the text embedding space, which in our setting is well captured by a single adaptive inclusive token. Introducing multiple inclusive tokens increases the expressive capacity of the text conditioning, but may also make the optimization more prone to overfitting spurious correlations present in the limited training data, such as background, clothing, or pose patterns shared across samples.}

\begin{table}[tb]
  \caption{Ablation study on the number of inclusive tokens (IT).}
  \centering
  \renewcommand\arraystretch{1.2}
    \setlength{\tabcolsep}{1.6mm}{
  \begin{tabular}{c | ccc}
        \hline
         \textbf{No. of IT} & \textbf{Gender} $\mathcal{D_{KL}}$ & \textbf{Race} $\mathcal{D_{KL}}$ & \textbf{Age} $\mathcal{D_{KL}}$ \\
         \hline
        1 (Ours) & \major{\textbf{0.1290}} & \major{0.3488} & \major{\textbf{0.2135}} \\
        2 & \major{0.2074} & \major{\textbf{0.3476}} & \major{0.2472} \\
        3 & \major{0.1981} & \major{0.4297} & \major{0.2423} \\
        \hline
    \end{tabular}
    }
    \label{tab:no}
\end{table}

\begin{table}[tb]
  \caption{Ablation study on the number of training samples for each attribute-occupation.}
  \centering
  \renewcommand\arraystretch{1.2}
    \setlength{\tabcolsep}{1.8mm}{
  \begin{tabular}{c | ccc}
        \hline
         \textbf{Number} & \textbf{Gender} $\mathcal{D_{KL} }$ & \textbf{Race} $\mathcal{D_{KL} }$ & \textbf{Age} $\mathcal{D_{KL} }$ \\
         \hline
        20 & \major{0.1573} & \major{0.3509} & \major{\textbf{0.2066}} \\
        10 & \major{\textbf{0.1290}} & \major{\textbf{0.3488}} & \major{0.2135} \\
        5 & \major{0.1988} & \major{0.4259} & \major{0.2434} \\
        \hline
    \end{tabular}
    }
    \label{tab:sample}
\end{table}

\subsubsection{\minor{Training Data Analysis}}

\noindent \textbf{Number of Training Samples.}
We analyze the impact of the number of training samples on bias mitigation tasks, with results presented in Tab.~\ref{tab:sample}. 
\major{We further investigate the impact of the number of training samples per attribute–occupation combination, as reported in Table 11. Interestingly, the results do not exhibit a monotonic trend where more samples always lead to better performance. This phenomenon can be explained in the context of Textual Inversion (TI), where embeddings are learned to represent concepts from a limited number of examples. Prior work on TI~\citep{gal2022textual, kumari2022customdiffusion} has shown that TI is inherently a few-shot learning paradigm, and a small number of carefully chosen reference images are often sufficient to capture the semantic essence of a concept. Using too many samples may, in fact, degrade the quality of the learned embeddings, since the optimization can overfit to irrelevant or spurious correlations (\eg, background or clothing shared across training samples) rather than focusing on the target attribute itself. This finding aligns with the discussion in Supplementary Materials Section B of \citet{gal2022textual}, where the authors find that 5 images per object operate the best, while additional images harm editability. In our case, this implies that excessive samples per attribute–occupation combination could reduce the specificity of the adaptive inclusive token, thereby limiting its effectiveness in debiasing. Taken together, these findings highlight that, similar to TI, more training data is not always beneficial for learning functional text embeddings; instead, a moderate number of examples (\eg, 10 in our experiments) often provides the best trade-off between stability and robustness.}

\begin{table}[tb]
  \caption{Ablation study on image sources. Reported metrics are on gender bias.}
  \centering
  \renewcommand\arraystretch{1.2}
    {
  \begin{tabular}{c | ccc}
        \hline
         \textbf{Image Source} & $\mathcal{D_{KL} \downarrow}$ & FID $\downarrow$ & CLIP $\uparrow$ \\
         \hline
        SD1.5 (888) & \major{\textbf{0.1520}} & \major{\textbf{268.72}} & \major{0.2789} \\
        Online (888) & \major{0.2386} & \major{273.06} & \major{\textbf{0.2815}} \\
        \hline
    \end{tabular}
    }
    \label{tab:source}
\end{table}

\noindent \textbf{Image Sources.}
We examine the impact of different image sources on bias mitigation performance. Due to the difficulty of obtaining real images for certain attribute-occupation combinations, this study is limited to gender bias.
To construct a real-image dataset, we use \texttt{clip-retrieval}~\citep{beaumont-2022-clip-retrieval} to retrieve 20 images per attribute-occupation combination from the LAION-400M dataset~\citep{schuhmann2021laion}. The retrieved images are manually screened to retain only those that correctly match the attribute classes and occupation-related visual concepts. To ensure balanced training data, we maintain an equal number of images for each gender within each occupation category, resulting in a total of 888 real images in the training set. 
For a fair comparison, we generate an equal number (888) of synthetic images using SD1.5, ensuring that the number of samples per attribute-occupation combination matches that of the real-image dataset. The quantitative results, presented in Tab.~\ref{tab:source}, indicate that training the adaptive inclusive token with synthetic images generated by the same SD model yields better inclusiveness and image quality. 
\major{As for text–image alignment, it is reasonable that models trained on real images outperform those trained on synthetic data, since CLIP itself was trained on large-scale natural image–text pairs, and embeddings derived from real images are therefore more consistent with its learned representation space.}

\section{Limitations}
\label{sec:limitation}

\minor{Despite its design to enable adaptive and concept-aware bias mitigation, AITTI remains subject to several fundamental limitations inherent to fairness-oriented text-to-image generation. These limitations arise from (i) the ambiguity between fairness and factual correctness for certain concepts, (ii) the dependence on sufficient attribute coverage in the training data, (iii) the difficulty of jointly mitigating multiple bias types within a single model, and (iv) the reliance on an imperfect attribute classifier for data preparation and evaluation. We discuss each aspect below.}

\minor{First, our method, like prior approaches, focuses on achieving an even attribute distribution and is therefore limited to neutral concepts, where all attribute classes are factually valid. For non-neutral concepts, debiasing may lead to factual inaccuracies, such as generating a Black individual as ``the pope'' when using a $\mathord{<}$race-inclusive$\mathord{>}$ token. Addressing these inaccuracies would require an additional fact-checking mechanism. Nevertheless, from an inclusiveness and creative perspective, such outcomes may not always be undesirable. The trade-off between accuracy and fairness remains a complex challenge, necessitating ongoing discussions among researchers and policymakers.}

\minor{Second, like most data-driven fairness approaches, our method cannot effectively mitigate biases for attribute categories that are completely absent from the training distribution, as the model lacks exposure to their visual or semantic representations. To reduce this risk in our experiments, we focus on attribute categories that are widely adopted in the fairness literature and sufficiently represented in benchmark datasets. At the same time, we note that our method does not rely on a fixed, pre-defined label space or per-class token training, unlike prior works such as ITI-GEN~\citep{zhang2023itigen}. As a result, incorporating new attribute categories does not require redesigning the framework or introducing additional class-specific components. This design makes our approach more amenable to extension as new categories become available, although its effectiveness ultimately still depends on data coverage.}

\minor{Third, the current framework does not support joint mitigation of multiple bias types using a single adaptive mapping network. Similar to most existing approaches, we train a separate bias-specific mapping for each sensitive attribute. This design reflects the difficulty of learning disentangled representations when multiple attributes co-occur with highly imbalanced distributions. Enabling unified multi-bias mitigation within a single model is a promising but challenging direction for future work.}

\minor{Finally, our approach relies on CLIP for both training data filtering and fairness evaluation, which introduces a potential source of bias propagation. This risk is inherent to CLIP-based pipelines and is shared by many recent text-to-image fairness studies. To partially mitigate this issue, we manually screened the training data to remove clearly mismatched or low-quality samples. Moreover, while CLIP does not provide perfect attribute predictions, particularly for challenging attributes such as age, where classification accuracy is known to be lower than for other attributes, it offers a consistent evaluation protocol across methods. As a result, although CLIP-related biases may affect absolute metric values, we expect the relative comparisons under the same evaluation setup to remain meaningful.}

\section{Conclusion}
\label{sec:conclusion}

In this work, we investigate the challenges of prompt tuning-based bias mitigation approaches and identify key limitations in their generalization across different dominant attribute classes. To address these issues, we propose learning a concept-specific inclusive token through an adaptive mapping network, rather than relying on a fixed token. Additionally, to mitigate concept drift, we introduce an anchor loss, which constrains the impact of the adaptive inclusive token on the final outputs. As a result, our method significantly enhances inclusiveness in T2I generation, demonstrating generalizability to unseen concepts and scenarios while effectively mitigating multiple biases simultaneously. Furthermore, our approach is model-agnostic, making it applicable to various T2I models to improve fairness in generated outputs.
We believe that fostering inclusivity in generative models is a collective responsibility that can contribute to a world where individuals, regardless of background or identity, are represented fairly and respectfully.

\noindent
\textbf{Data Availability.}
\textcolor{black}{The datasets that support the findings of this study are all publicly available online and available for research purposes.
The training data are generated using Huggingface diffusers SD1.5 implementation that are available at \url{https://github.com/huggingface/diffusers}.}

\bibliographystyle{spbasic}      
\bibliography{ref}   

\begin{thebibliography}{45}
\providecommand{\natexlab}[1]{#1}
\providecommand{\url}[1]{{#1}}
\providecommand{\urlprefix}{URL }
\expandafter\ifx\csname urlstyle\endcsname\relax
  \providecommand{\doi}[1]{DOI~\discretionary{}{}{}#1}\else
  \providecommand{\doi}{DOI~\discretionary{}{}{}\begingroup \urlstyle{rm}\Url}\fi
\providecommand{\eprint}[2][]{\url{#2}}

\bibitem[{Bansal et~al.(2022)Bansal, Yin, Monajatipoor, and Chang}]{bansal2022howwell}
Bansal H, Yin D, Monajatipoor M, Chang KW (2022) How well can text-to-image generative models understand ethical natural language interventions?

\bibitem[{Beaumont(2022)}]{beaumont-2022-clip-retrieval}
Beaumont R (2022) Clip retrieval: Easily compute clip embeddings and build a clip retrieval system with them. \url{https://github.com/rom1504/clip-retrieval}

\bibitem[{Bianchi et~al.(2023)Bianchi, Kalluri, Durmus, Ladhak, Cheng, Nozza, Hashimoto, Jurafsky, Zou, and Caliskan}]{bianchi2023easily}
Bianchi F, Kalluri P, Durmus E, Ladhak F, Cheng M, Nozza D, Hashimoto T, Jurafsky D, Zou J, Caliskan A (2023) Easily accessible text-to-image generation amplifies demographic stereotypes at large scale. In: ACM Conference on Fairness, Accountability, and Transparency

\bibitem[{BlackForestLabs(2023)}]{flux_github}
BlackForestLabs (2023) {FLUX}. \url{https://github.com/black-forest-labs/flux}

\bibitem[{Chinchure et~al.(2023)Chinchure, Shukla, Bhatt, Salij, Hosanagar, Sigal, and Turk}]{chinchure2023tibet}
Chinchure A, Shukla P, Bhatt G, Salij K, Hosanagar K, Sigal L, Turk M (2023) {TIBET}: Identifying and evaluating biases in text-to-image generative models. arXiv preprint arXiv: 231201261

\bibitem[{Choi et~al.(2024)Choi, Park, Kim, Lee, and Park}]{choi2024switchmechanism}
Choi Y, Park J, Kim H, Lee J, Park S (2024) Fair sampling in diffusion models through switching mechanism. In: Proceedings of the AAAI Conference on Artificial Intelligence

\bibitem[{Deng et~al.(2009)Deng, Dong, Socher, Li, Li, and Fei-Fei}]{imagenet}
Deng J, Dong W, Socher R, Li LJ, Li K, Fei-Fei L (2009) Imagenet: A large-scale hierarchical image database. In: Proceedings of the IEEE/CVF Conference on Computer Vision and Pattern Recognition (CVPR)

\bibitem[{Deng et~al.(2020)Deng, Guo, Ververas, Kotsia, and Zafeiriou}]{deng2020retinaface}
Deng J, Guo J, Ververas E, Kotsia I, Zafeiriou S (2020) {RetinaFace}: Single-shot multi-level face localisation in the wild. In: Proceedings of the IEEE/CVF Conference on Computer Vision and Pattern Recognition (CVPR)

\bibitem[{D'Inc\`a et~al.(2024)D'Inc\`a, Peruzzo, Mancini, Xu, Goel, Xu, Wang, Shi, and Sebe}]{openbias}
D'Inc\`a M, Peruzzo E, Mancini M, Xu D, Goel V, Xu X, Wang Z, Shi H, Sebe N (2024) Openbias: Open-set bias detection in text-to-image generative models. In: Proceedings of the IEEE/CVF Conference on Computer Vision and Pattern Recognition (CVPR)

\bibitem[{Esser et~al.(2024)Esser, Kulal, Blattmann, Entezari, M{\"u}ller, Saini, Levi, Lorenz, Sauer, Boesel, Podell, Dockhorn, English, and Rombach}]{sd3}
Esser P, Kulal S, Blattmann A, Entezari R, M{\"u}ller J, Saini H, Levi Y, Lorenz D, Sauer A, Boesel F, Podell D, Dockhorn T, English Z, Rombach R (2024) Scaling rectified flow transformers for high-resolution image synthesis. In: ICML

\bibitem[{Friedrich et~al.(2023)Friedrich, Brack, Struppek, Hintersdorf, Schramowski, Luccioni, and Kersting}]{friedrich2023fairdiffusion}
Friedrich F, Brack M, Struppek L, Hintersdorf D, Schramowski P, Luccioni S, Kersting K (2023) {Fair Diffusion}: Instructing text-to-image generation models on fairness. arXiv preprint arXiv: 230210893

\bibitem[{Gal et~al.(2022)Gal, Alaluf, Atzmon, Patashnik, Bermano, Chechik, and Cohen-Or}]{gal2022textual}
Gal R, Alaluf Y, Atzmon Y, Patashnik O, Bermano AH, Chechik G, Cohen-Or D (2022) An image is worth one word: Personalizing text-to-image generation using textual inversion. In: Proceedings of International Conference on Learning Representations (ICLR)

\bibitem[{Gandikota et~al.(2024)Gandikota, Orgad, Belinkov, Materzyńska, and Bau}]{gandikota2023unified}
Gandikota R, Orgad H, Belinkov Y, Materzyńska J, Bau D (2024) Unified concept editing in diffusion models. In: Proceedings of the IEEE/CVF Winter Conference on Applications of Computer Vision (WACV)

\bibitem[{Ghosh and Caliskan(2023)}]{ghosh2023person}
Ghosh S, Caliskan A (2023) 'person' == light-skinned, western man, and sexualization of women of color: Stereotypes in stable diffusion

\bibitem[{Greenwald et~al.(1998)Greenwald, McGhee, and Schwartz}]{Greenwald1998IAT}
Greenwald AG, McGhee DE, Schwartz JLK (1998) Measuring individual differences in implicit cognition: the implicit association test. Journal of personality and social psychology 74 6:1464--80

\bibitem[{Heusel et~al.(2017)Heusel, Ramsauer, Unterthiner, Nessler, and Hochreiter}]{heusel2017fid}
Heusel M, Ramsauer H, Unterthiner T, Nessler B, Hochreiter S (2017) Gans trained by a two time-scale update rule converge to a local nash equilibrium. In: Proceedings of Advances in Neural Information Processing Systems (NeurIPS)

\bibitem[{Ho and Salimans(2021)}]{ho2021classifier}
Ho J, Salimans T (2021) Classifier-free diffusion guidance. In: NeurIPS 2021 Workshop on Deep Generative Models and Downstream Applications

\bibitem[{Jha et~al.(2024)Jha, Prabhakaran, Denton, Laszlo, Dave, Qadri, Reddy, and Dev}]{jha2024surface}
Jha A, Prabhakaran V, Denton R, Laszlo S, Dave S, Qadri R, Reddy CK, Dev S (2024) Beyond the surface: A global-scale analysis of visual stereotypes in text-to-image generation. arXiv preprint arXiv: 240106310

\bibitem[{Jiang et~al.(2024)Jiang, Li, Zhang, Cai, and Yue}]{jiang2024debiasdiff}
Jiang Y, Li W, Zhang Y, Cai M, Yue X (2024) Debiasdiff: Debiasing text-to-image diffusion models with self-discovering latent attribute directions. arXiv preprint arXiv:241218810

\bibitem[{Karkkainen and Joo(2021)}]{karkkainen2021fairface}
Karkkainen K, Joo J (2021) {FairFace}: Face attribute dataset for balanced race, gender, and age for bias measurement and mitigation. In: Proceedings of the IEEE/CVF Winter Conference on Applications of Computer Vision (WACV)

\bibitem[{Kim et~al.(2023)Kim, Kim, Shin, and Yoon}]{kim2023destereotyping}
Kim E, Kim S, Shin C, Yoon S (2023) De-stereotyping text-to-image models through prompt tuning. In: ICML Workshop

\bibitem[{Kumari et~al.(2023)Kumari, Zhang, Zhang, Shechtman, and Zhu}]{kumari2022customdiffusion}
Kumari N, Zhang B, Zhang R, Shechtman E, Zhu JY (2023) Multi-concept customization of text-to-image diffusion. In: Proceedings of the IEEE/CVF Conference on Computer Vision and Pattern Recognition (CVPR)

\bibitem[{Li et~al.(2023)Li, Hu, Zhang, Zheng, Zhang, and Wang}]{li2023fairmapping}
Li J, Hu L, Zhang J, Zheng T, Zhang H, Wang D (2023) Fair text-to-image diffusion via fair mapping. arXiv preprint arXiv: 231117695

\bibitem[{Loshchilov and Hutter(2019)}]{loshchilov2018decoupled}
Loshchilov I, Hutter F (2019) Decoupled weight decay regularization. In: Proceedings of International Conference on Learning Representations (ICLR)

\bibitem[{Lyu et~al.(2025)Lyu, Yang, Niu, Jiang, and Lo}]{lyu2025existing}
Lyu Y, Yang Z, Niu Y, Jiang J, Lo D (2025) Do existing testing tools really uncover gender bias in text-to-image models? arXiv preprint arXiv:250115775

\bibitem[{Orgad et~al.(2023)Orgad, Kawar, and Belinkov}]{orgad2023time}
Orgad H, Kawar B, Belinkov Y (2023) {TIME}: Editing implicit assumptions in text-to-image diffusion models. In: Proceedings of the IEEE/CVF International Conference on Computer Vision (ICCV)

\bibitem[{Parihar et~al.(2024)Parihar, Bhat, Basu, Mallick, Kundu, and Babu}]{parihar2024balancing}
Parihar R, Bhat A, Basu A, Mallick S, Kundu JN, Babu RV (2024) Balancing act: Distribution-guided debiasing in diffusion models. In: Proceedings of the IEEE/CVF Conference on Computer Vision and Pattern Recognition (CVPR)

\bibitem[{Podell et~al.(2024)Podell, English, Lacey, Blattmann, Dockhorn, M{\"u}ller, Penna, and Rombach}]{podell2023sdxl}
Podell D, English Z, Lacey K, Blattmann A, Dockhorn T, M{\"u}ller J, Penna J, Rombach R (2024) {SDXL}: Improving latent diffusion models for high-resolution image synthesis. In: Proceedings of International Conference on Learning Representations (ICLR)

\bibitem[{Radford et~al.(2021)Radford, Kim, Hallacy, Ramesh, Goh, Agarwal, Sastry, Askell, Mishkin, Clark et~al.}]{radford2021clip}
Radford A, Kim JW, Hallacy C, Ramesh A, Goh G, Agarwal S, Sastry G, Askell A, Mishkin P, Clark J, et~al. (2021) Learning transferable visual models from natural language supervision. In: Proceedings of International Conference on Machine Learning (ICML)

\bibitem[{Rombach et~al.(2022)Rombach, Blattmann, Lorenz, Esser, and Ommer}]{Rombach2022stable}
Rombach R, Blattmann A, Lorenz D, Esser P, Ommer B (2022) High-resolution image synthesis with latent diffusion models. In: Proceedings of the IEEE/CVF Conference on Computer Vision and Pattern Recognition (CVPR)

\bibitem[{Ronneberger et~al.(2015)Ronneberger, Fischer, and Brox}]{unet}
Ronneberger O, Fischer P, Brox T (2015) {U-Net}: Convolutional networks for biomedical image segmentation

\bibitem[{Ruiz et~al.(2023)Ruiz, Li, Jampani, Pritch, Rubinstein, and Aberman}]{ruiz2022dreambooth}
Ruiz N, Li Y, Jampani V, Pritch Y, Rubinstein M, Aberman K (2023) {DreamBooth}: Fine tuning text-to-image diffusion models for subject-driven generation. In: Proceedings of the IEEE/CVF Conference on Computer Vision and Pattern Recognition (CVPR)

\bibitem[{Runway(2023)}]{runwayresearch2023mitigating}
Runway R (2023) Mitigating stereotypical biases in text to image generative systems. arXiv preprint arXiv: 231006904

\bibitem[{Schuhmann et~al.(2021)Schuhmann, Vencu, Beaumont, Kaczmarczyk, Mullis, Katta, Coombes, Jitsev, and Komatsuzaki}]{schuhmann2021laion}
Schuhmann C, Vencu R, Beaumont R, Kaczmarczyk R, Mullis C, Katta A, Coombes T, Jitsev J, Komatsuzaki A (2021) {LAION-400M}: Open dataset of clip-filtered 400 million image-text pairs. In: NeurIPS Workshop

\bibitem[{Shen et~al.(2024)Shen, Du, Pang, Lin, Wong, and Kankanhalli}]{shen2023finetuning}
Shen X, Du C, Pang T, Lin M, Wong Y, Kankanhalli M (2024) Finetuning text-to-image diffusion models for fairness. In: Proceedings of International Conference on Learning Representations (ICLR)

\bibitem[{Sun et~al.(2024)Sun, Wei, Sun, Suh, Shen, and Yang}]{smiling}
Sun L, Wei M, Sun Y, Suh YJ, Shen L, Yang S (2024) Smiling women pitching down: auditing representational and presentational gender biases in image-generative ai. In: Journal of Computer-Mediated Communication, vol~29, p zmad045

\bibitem[{Teo et~al.(2023)Teo, Abdollahzadeh, and Cheung}]{teo2023measuring}
Teo C, Abdollahzadeh M, Cheung NMM (2023) On measuring fairness in generative models. In: Proceedings of Advances in Neural Information Processing Systems (NeurIPS)

\bibitem[{Teo et~al.(2024)Teo, Abdollahzadeh, Ma, and Cheung}]{teo2024fairqueue}
Teo C, Abdollahzadeh M, Ma X, Cheung NMM (2024) Fairqueue: Rethinking prompt learning for fair text-to-image generation. In: Proceedings of Advances in Neural Information Processing Systems (NeurIPS)

\bibitem[{Wang et~al.(2023)Wang, Liu, Di, Liu, and Wang}]{wang2023t2iat}
Wang J, Liu XG, Di Z, Liu Y, Wang X (2023) {T2IAT}: Measuring valence and stereotypical biases in text-to-image generation. In: Findings of the Association for Computational Linguistics

\bibitem[{Wang et~al.(2024)Wang, Bai, Huang, Wan, Yuan, Qiu, Peng, and Lyu}]{wang2024new}
Wang W, Bai H, Huang Jt, Wan Y, Yuan Y, Qiu H, Peng N, Lyu MR (2024) New job, new gender? measuring the social bias in image generation models. In: Proceedings of the ACM International Conference on Multimedia (ACM MM)

\bibitem[{Yesiltepe et~al.(2024)Yesiltepe, Akdemir, and Yanardag}]{yesiltepe2024mist}
Yesiltepe H, Akdemir K, Yanardag P (2024) Mist: Mitigating intersectional bias with disentangled cross-attention editing in text-to-image diffusion models. arXiv preprint arXiv:240319738

\bibitem[{Zhang et~al.(2023)Zhang, Chen, Chai, Wu, Lagun, Beeler, and De~la Torre}]{zhang2023itigen}
Zhang C, Chen X, Chai S, Wu CH, Lagun D, Beeler T, De~la Torre F (2023) {ITI-GEN}: Inclusive text-to-image generation. In: Proceedings of the IEEE/CVF International Conference on Computer Vision (ICCV)

\bibitem[{Zhou et~al.(2024)Zhou, Gao, Zhao, Yao, and Wei}]{zhou2024association}
Zhou J, Gao J, Zhao X, Yao X, Wei X (2024) Association of objects may engender stereotypes: Mitigating association-engendered stereotypes in text-to-image generation. In: Proceedings of Advances in Neural Information Processing Systems (NeurIPS)

\bibitem[{Zhou et~al.(2022{\natexlab{a}})Zhou, Yang, Loy, and Liu}]{zhou2022cocoop}
Zhou K, Yang J, Loy CC, Liu Z (2022{\natexlab{a}}) Conditional prompt learning for vision-language models. In: Proceedings of the IEEE/CVF Conference on Computer Vision and Pattern Recognition (CVPR)

\bibitem[{Zhou et~al.(2022{\natexlab{b}})Zhou, Yang, Loy, and Liu}]{zhou2022coop}
Zhou K, Yang J, Loy CC, Liu Z (2022{\natexlab{b}}) Learning to prompt for vision-language models. In: International Journal of Computer Vision (IJCV)

\end{thebibliography}

\newpage
\appendix
\section{Occupation Lists}
\label{sec:occ list}

The training occupations are:
\texttt{["construction worker", "mechanic", "firefighter", "police officer", "engineer", "pilot", "computer programmer", "ceo", "surgeon", "truck driver", "electrician", "scientist", "nurse", "teacher", "secretary", "librarian", "receptionist", "administrative assistant", "childcare worker", "social worker", "dental hygienist", "therapist", "pathologist", "nutritionist"]}

\noindent
The testing occupations are:
\texttt{["doctor", "chief", "farmer", "architect", "software developer", "ballet dancer", "yoga instructor", "cosmetologist", "fashion designer", "flight attendant", "artist", "musician", "writer", "photographer", "philosopher", "athlete", "lawyer", "politician", "journalist", "barista", "detective", "security guard", "professor", "sports coach"]}

\section{ImageNet Prompt Templates}
\label{sec:prompttemp}

The prompt templates for training are:

\noindent
\texttt{[
    "a photo of a \{attribute\} \{occupation\}", \\
    "a rendering of a \{attribute\} \{occupation\}", \\
    "a cropped photo of the \{attribute\} \{occupation\}", \\
    "the photo of a \{attribute\} \{occupation\}", \\
    "a photo of a clean \{attribute\} \{occupation\}", \\
    "a photo of a dirty \{attribute\} \{occupation\}", \\
    "a dark photo of the \{attribute\} \{occupation\}", \\
    "a photo of my \{attribute\} \{occupation\}", \\
    "a photo of the cool \{attribute\} \{occupation\}", \\
    "a close-up photo of a \{attribute\} \{occupation\}", \\
    "a bright photo of the \{attribute\} \{occupation\}", \\
    "a cropped photo of a \{attribute\} \{occupation\}", \\
    "a photo of the \{attribute\} \{occupation\}", \\
    "a good photo of the \{attribute\} \{occupation\}", \\
    "a photo of one \{attribute\} \{occupation\}", \\
    "a close-up photo of the \{attribute\} \{occupation\}", \\
    "a rendition of the \{attribute\} \{occupation\}", \\
    "a photo of the clean \{attribute\} \{occupation\}", \\
    "a rendition of a \{attribute\} \{occupation\}", \\
    "a photo of a nice \{attribute\} \{occupation\}", \\
    "a good photo of a \{attribute\} \{occupation\}", \\
    "a photo of the nice \{attribute\} \{occupation\}", \\
    "a photo of the small \{attribute\} \{occupation\}", \\
    "a photo of the weird \{attribute\} \{occupation\}", \\
    "a photo of the large \{attribute\} \{occupation\}", \\
    "a photo of a cool \{attribute\} \{occupation\}", \\
    "a photo of a small \{attribute\} \{occupation\}" \\
    ]}

\begin{figure}[tb]
    \centering
    \includegraphics[width=0.98\linewidth]{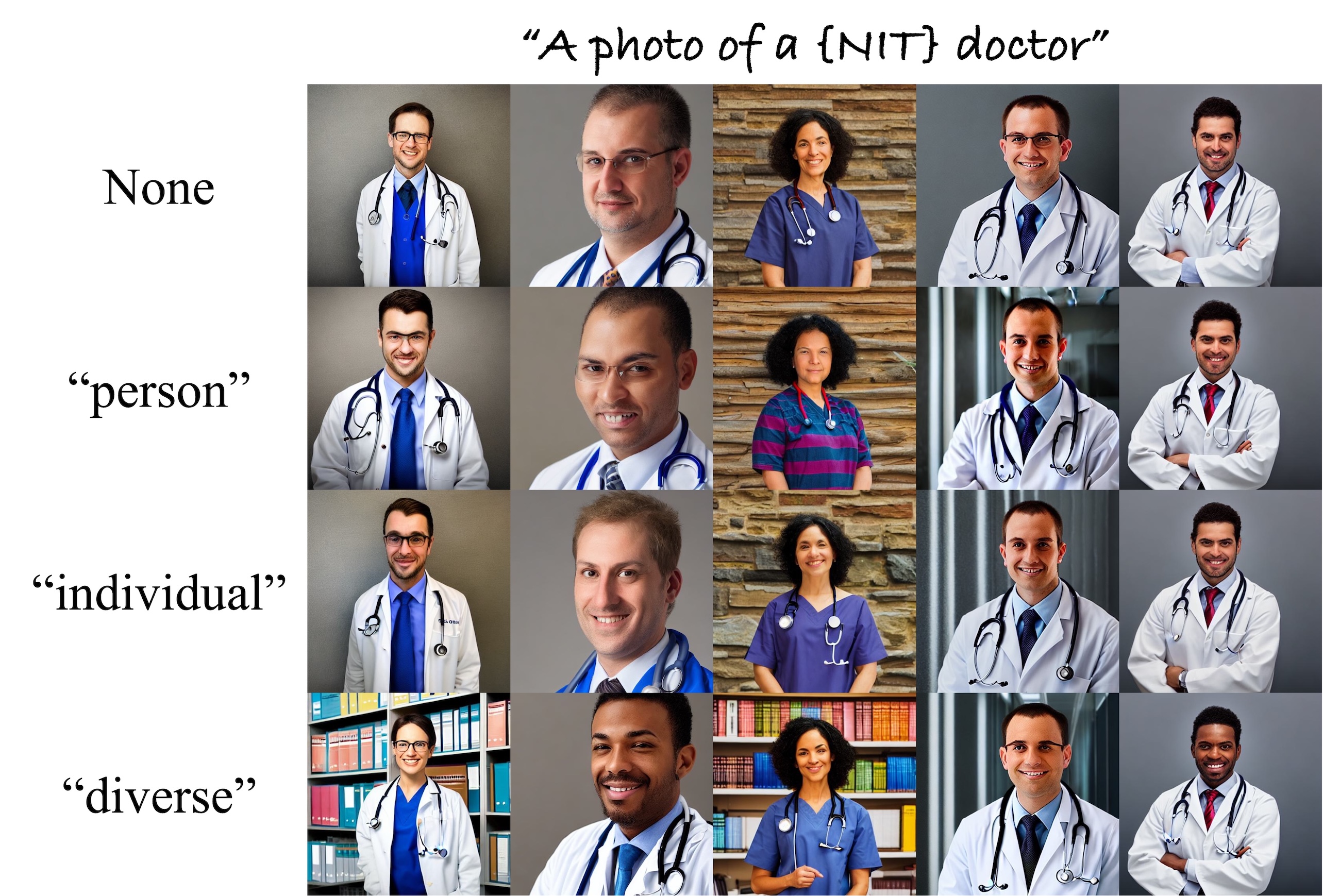}
    \caption{Visual effects of different natural inclusive tokens in an occupation-related prompt.}
    \label{fig:nit}
\end{figure}

\begin{table}[tb]
  \caption{Effect of NIT in gender bias mitigation. The testing prompt is ``A photo of a [NIT] \{occupation\}'' and the testing occupations consist of 10 unseen occupations (5 from each gender domination group).}
  \centering
  \renewcommand\arraystretch{1.2}
    \setlength{\tabcolsep}{1.3mm}{
  \begin{tabular}{c | ccc}
        \hline
         \textbf{\makecell{Natural Inclusive \\Token (NIT)}} & \textbf{Gender} $\mathcal{D_{KL}}$ & FID & CLIP \\
         \hline
        None & 0.4783 & 287.55 & 0.2909 \\
        \hline
        ``individual'' & 0.4222 & 287.27 & \textbf{0.2960} \\
        ``person'' & 0.4021 & \textbf{282.76} & 0.2936 \\
        ``diverse'' & \textbf{0.3122} & 299.14 & 0.2875 \\
        \hline
    \end{tabular}
    }
    \label{tab:nit}
\end{table}

\section{Natural Inclusive Token}
\label{sec:nit-init}

\textcolor{black}{
We observe the existence of natural inclusive tokens (NITs), which we define as tokens that, when incorporated into occupation-related prompts (\eg, ``A photo of a [NIT] {occupation}''), promote a fairer attribute distribution without significantly altering the structure or semantics of the generated images. The visual effects of NIT-based prompts are illustrated in Fig.~\ref{fig:nit}. To explore their effectiveness, we evaluate several NIT candidates on gender bias mitigation across 10 unseen occupations from different dominant gender groups, following the same setup as in Tab.~\ref{tab:chief_comparisons-a}. As shown in Tab.~\ref{tab:nit}, NITs can partially reduce distribution discrepancies, though they are not as effective as the learned adaptive inclusive tokens.}

\section{More Qualitative Results}
\label{sec:qualitative}
\begin{figure*}[h!]
    \centering
    \includegraphics[width=0.8\textwidth]{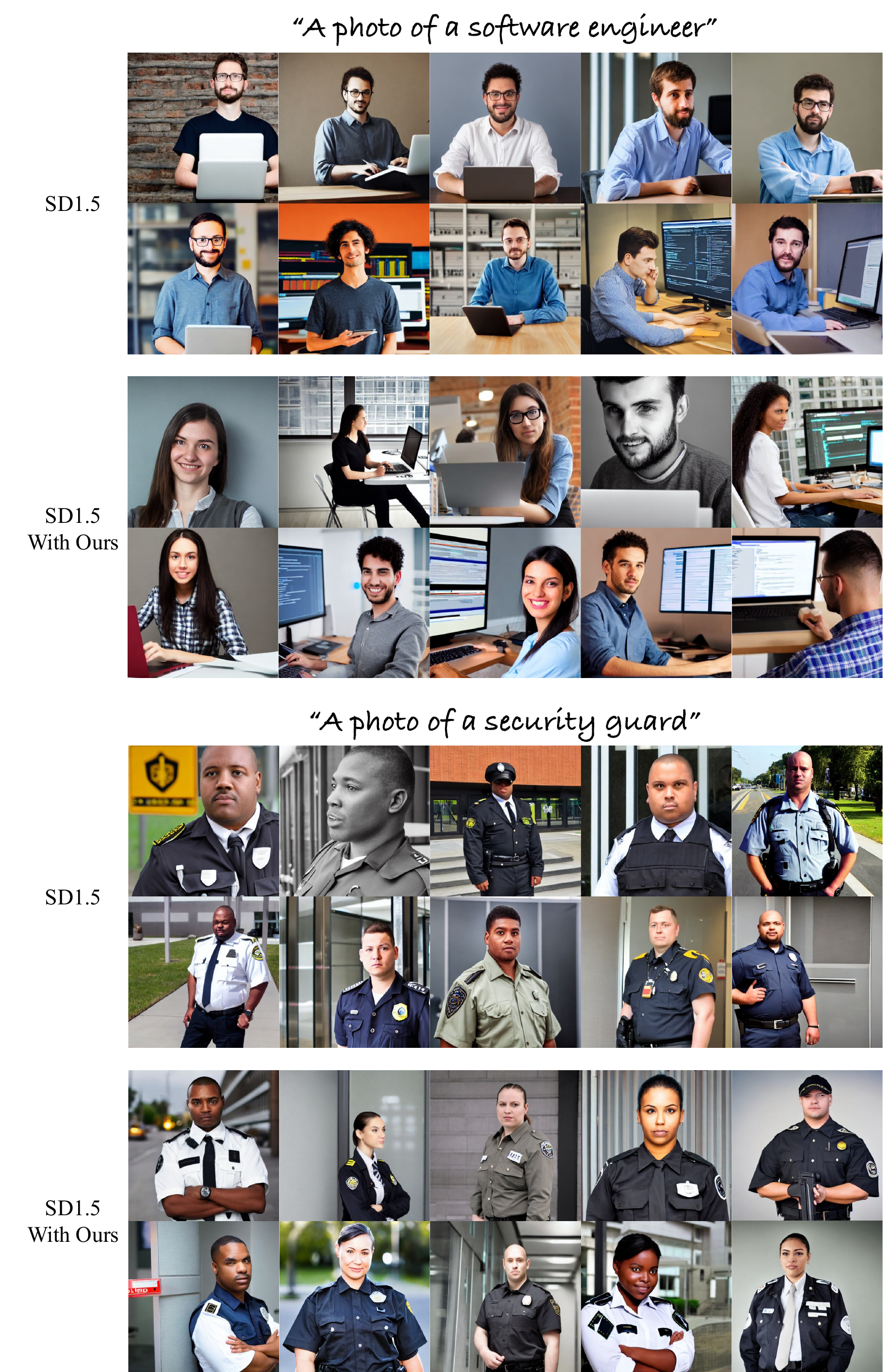}
    \caption{Qualitative results of gender bias mitigation on SD1.5. All images are generated with the same random seed.}
    \label{fig:sd15-gender}
\end{figure*}

\begin{figure*}[h!]
    \centering
    \includegraphics[width=0.8\textwidth]{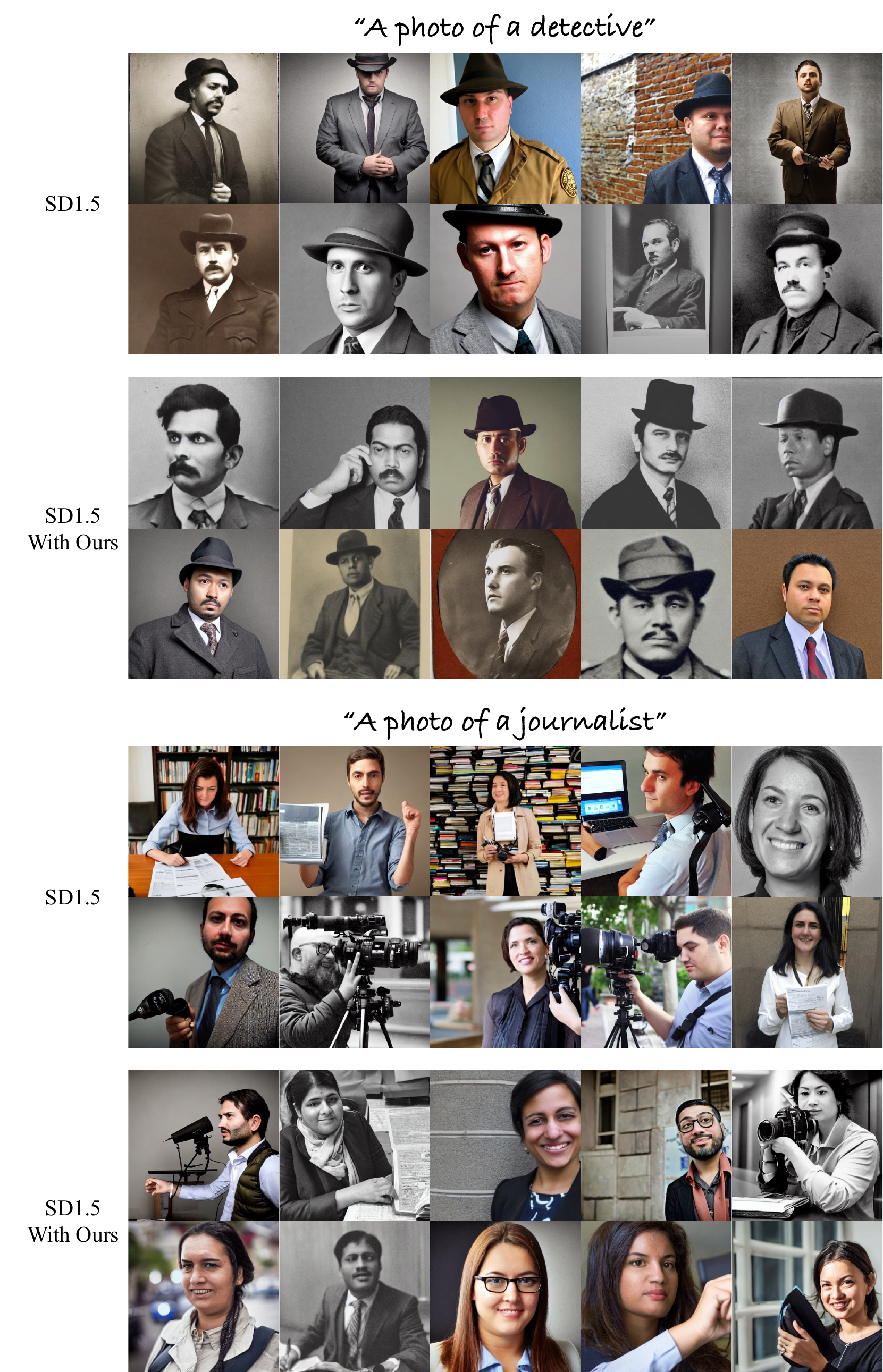}
    \caption{Qualitative results of race bias mitigation on SD1.5. All images are generated with the same random seed.}
    \label{fig:sd15-race}
\end{figure*}

\begin{figure*}[h!]
    \centering
    \includegraphics[width=0.8\textwidth]{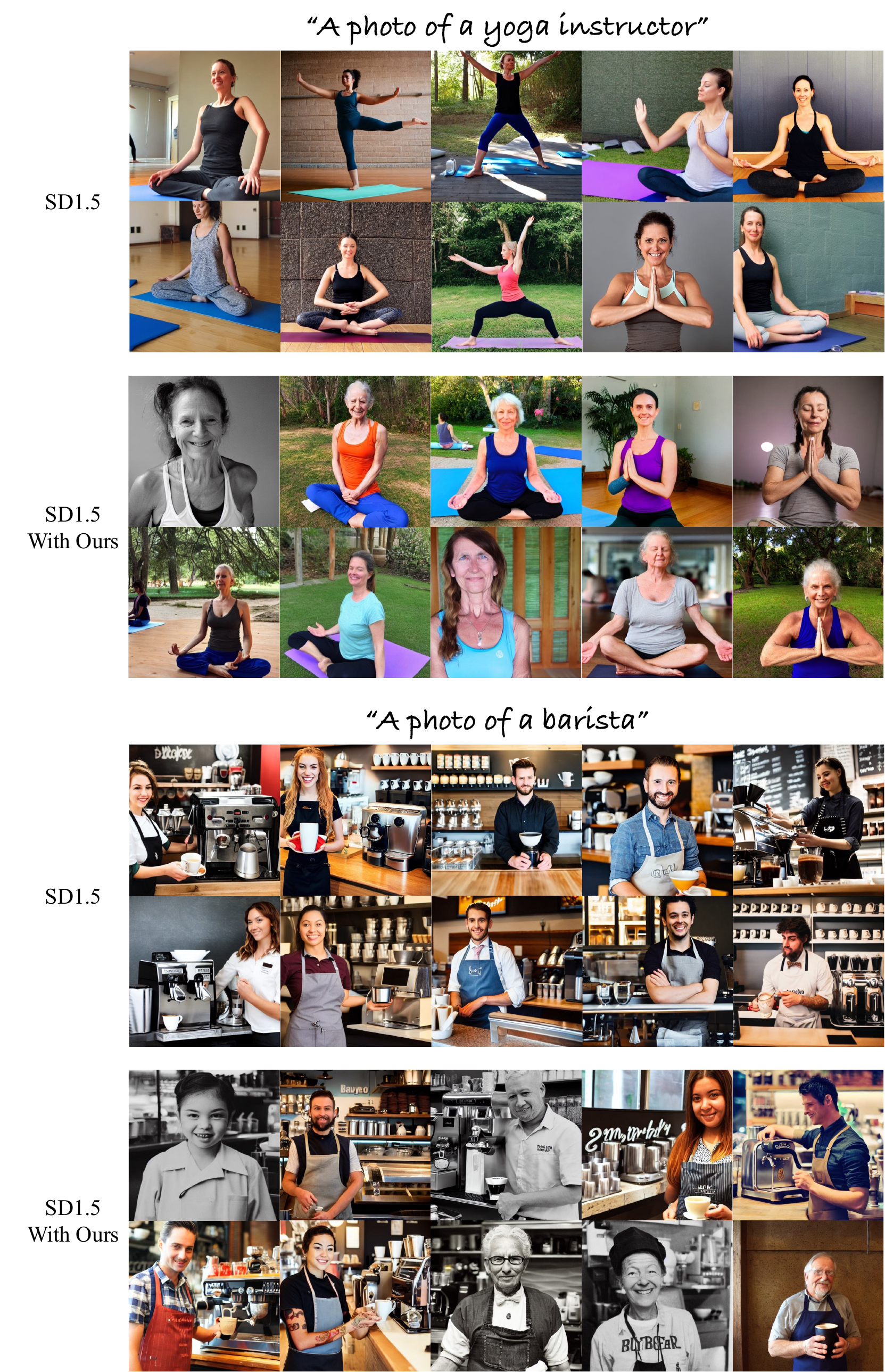}
    \caption{Qualitative results of age bias mitigation on SD1.5. All images are generated with the same random seed.}
    \label{fig:sd15-age}
\end{figure*}

\vspace{-8pt}
More qualitative results of applying our adaptive inclusive token on the SD1.5 model are presented in Figs.~\ref{fig:sd15-gender}, \ref{fig:sd15-race}, \ref{fig:sd15-age}.

\end{document}